\documentclass[11pt,a4paper]{article}

\usepackage[utf8]{inputenc}
\usepackage[T1]{fontenc}
\usepackage{lmodern}
\usepackage[margin=1in]{geometry}
\usepackage{graphicx}
\usepackage{booktabs}
\usepackage{amsmath,amssymb}
\usepackage{siunitx}
\usepackage{multirow}
\usepackage{xcolor}
\usepackage[font=small,labelfont=bf]{caption}
\usepackage{subcaption}
\usepackage[hidelinks]{hyperref}
\usepackage{cleveref}

\usepackage{placeins}        

\graphicspath{{figures/}}
\DeclareSIUnit{\pixel}{px}   

\newcommand{\net}{\ensuremath{\mathrm{net}}}
\newcommand{\asr}{ASR}
\newcommand{\dE}{\ensuremath{\Delta E_{00}}}
\DeclareMathOperator{\relu}{ReLU}
\DeclareMathOperator*{\Exp}{\mathbb{E}}

\title{\bfseries Printability-Constrained Adversarial Decals for Near-Nadir
Aerial Perception: Measured Ink Gamuts, Nested Realism Constraints,
and a Physical-World Bound}

\author{%
  Sandesh Shrestha\,$^{1}$ \and
  K.~T. Yasas Mahima\,$^{2}$ \and
  Asanka G. Perera\,$^{2,3,*}$
}

\date{}

\begin{document}
\maketitle

\begin{center}
\begin{minipage}{0.92\linewidth}
\footnotesize\raggedright
$^{1}$~Industrial Systems Engineering (ISE), Asian Institute of Technology
(AIT), Pathum Thani 12120, Thailand; \texttt{sandeshshrestha45@gmail.com};
ORCID 0000-0001-5298-4468\\[1pt]
$^{2}$~School of Engineering and Technology, University of New South Wales,
Canberra, ACT 2612, Australia; \texttt{yasas.mahima@unsw.edu.au};
ORCID 0000-0003-4975-9408\\[1pt]
$^{3}$~School of Engineering \& Digital Technologies, University of Southern
Queensland, Springfield Central, QLD 4300, Australia;
ORCID 0000-0003-4021-3943\\[1pt]
$^{*}$~Correspondence: \texttt{asanka.perera@unisq.edu.au}
\end{minipage}
\end{center}
\vspace{4pt}

\begin{abstract}
Adversarial patches for aerial perception are typically evaluated as digital composites, with printing left as an implementation detail. This study imposes three physical constraints during optimization rather than after it: the color range a particular printer can reproduce,
the size of the flat panel a vehicle offers, and the loss of fine detail incurred when the patch is imaged from altitude. The principal comparison isolates the ink set. Two patches share all seventeen recorded
optimization settings and differ only in the colors available to them. One is constrained to a uniform color cube; the other to a gamut measured by printing and scanning a 216-patch chart. Each was optimized at three seeds and evaluated against thirteen victim conditions, with every rate reported against a size-matched optimized control. The effect of the measured gamut is victim-dependent rather than uniform. Net attack success rises on three of six closed-set segmentation victims, and for these the seed ranges
of the two ink sets are disjoint: $+0.120$ on DeepLabv3-R101 and $+0.041$ on SegFormer-B0. The color-cube patch is consistently stronger on the open-vocabulary segmenter and on two of four detectors, though no detector exceeds a net of $+0.026$ under either ink set. The natural explanation is that a printable palette is simply less chromatic and lower
in frequency than a digital one. Eleven further patches test this account and it does not hold. Once cardinality is matched, a palette as chromatic as the cube attacks equally well. Cardinality itself shows no trend from three inks to thirty-two. Palettes matched on cardinality, lightness and chroma, and differing only in hue placement, span $0.035$ to $0.136$.
A physical evaluation with printed decals did not detect transfer; it bounds the transferred rate at $0.133$, which does not exclude the simulated value of $0.121$.
\end{abstract}


\section{Introduction}\label{sec:intro}
An adversarial patch attack on aerial perception implies a physical claim: that
an operator could apply a printed object to a vehicle and cause a downstream
model to fail. Most published evidence for this class of attack is digital.
Patches are optimised as unconstrained pixel arrays, composited into held-out
frames, and scored; the transition from composite to printed, photographed decal
is generally treated as an implementation step.

That transition imposes at least three constraints, each of which removes
degrees of freedom from the attacker:

\begin{itemize}
  \item A printer reproduces a bounded colour gamut rather than a colour space,
        so a patch optimised over $[0,1]^3$ will generally contain colours that
        cannot be reproduced.
  \item A decal is applied to a flat panel of bounded extent, so a patch larger
        than that panel corresponds to a composite onto a surface that does not
        exist on the vehicle.
  \item Imaging at altitude imposes a band limit, so texture above the physical
        channel cutoff cannot reach the sensor.
\end{itemize}

\noindent These constraints are commonly acknowledged in discussion sections and
less commonly incorporated into the optimisation. Non-printability scores, for
example, are widely reported, but typically against a generic colour set rather
than a measurement of the printer used \cite{sharif2016accessorize}. In this
study all three are made explicit, the gamut is measured rather than assumed,
and the physical experiment that the attack claim implies is attempted.

\paragraph{Contributions.}
\begin{enumerate}
  \item A controlled comparison of two patches identical in all seventeen
        recorded optimisation settings except the ink set, each optimised at
        three random seeds and scored against all thirteen victim conditions.
        Constraining the optimiser to a measured printer gamut does not
        uniformly reduce net attack success: it raises it on three of six
        closed-set segmentation victims with disjoint seed ranges, and lowers
        it on the open-vocabulary segmenter and on two of four detectors. The
        sign of the printability constraint is victim-dependent
        (\Cref{sec:res-gamut,sec:res-seeds}).
  \item A palette ablation of eleven further patches showing that neither the
        chromatic extent nor the cardinality of the ink set accounts for that
        effect, while palettes matched on both but differing in hue placement
        span three times the optimiser's between-seed variation. What a
        printer's gamut contributes is a particular set of colours rather than
        a smaller or less saturated one (\Cref{sec:res-h1}).
  \item Grad-CAM evidence \cite{selvaraju2017grad} that the mechanism is
        suppression of vehicle evidence beneath the decal rather than diversion
        of attribution toward it (\Cref{sec:res-cam}).
  \item Size-matched controls for every reported attack rate, with the
        observation that raw detection attack success is largely attributable to
        the presence of an applied object (\Cref{sec:res-det}).
  \item A physical-world evaluation reported as an upper confidence bound, with
        an explicit statement of the conditions under which the test is
        underpowered (\Cref{sec:res-phys}).
  \item A fiducial-free procedure for recovering effective altitude in physical
        captures from the known dimension of the photographed object
        (\Cref{sec:method-phys}).
\end{enumerate}

\FloatBarrier
\section{Related Work}\label{sec:related}

The field has been surveyed comprehensively: Wei et al.\ \cite{wei2024decade}
cover a decade of physical adversarial attacks across tasks and modalities,
Guesmi et al.\ \cite{guesmi2023survey} categorise attacks on camera-based
systems by deployment mechanism, and Li et al.\ \cite{li2025review} trace the
progression from planar patches to three-dimensional camouflage. The review
below is therefore selective, covering the strands this study builds on rather
than the field as a whole.

\subsection{Adversarial patches and dense prediction}
Localised, universal perturbations were introduced as adversarial patches by
Brown et al.\ \cite{brown2017patch}, who showed that a bounded image region can
dominate a classifier's prediction irrespective of scene content. Extension to
dense prediction followed. Xie et al.\ \cite{xie2017adversarial} demonstrated
adversarial examples against both semantic segmentation and object detection,
and Arnab et al.\ \cite{arnab2018robustness} characterised the robustness of
segmentation architectures to such perturbations, reporting that
multi-scale processing and dilated convolutions affect vulnerability. Attacks
targeting vision transformers have since been reported
\cite{fu2022patchfool}, alongside broader analyses of transformer robustness
\cite{bhojanapalli2021understanding,shao2021adversarial}. These motivate the
inclusion of both convolutional and transformer-based victims in
\Cref{sec:setup}.

\subsection{Physical-world attacks and printability}
Sharif et al.\ \cite{sharif2016accessorize} attacked face recognition using
printed eyeglass frames and introduced a non-printability score penalising
colours outside a reference set; that score is computed against a fixed
reference palette rather than a measurement of the printer used, which is the
practice this study departs from. Athalye et al.\
\cite{athalye2018synthesizing} formalised Expectation over Transformation (EOT),
optimising perturbations to be robust to a distribution of imaging
transformations. Thys et al.\ \cite{thys2019fooling} printed a patch that evaded
a person detector at ground level, and Eykholt et al.\ \cite{eykholt2018robust}
applied printed perturbations to road signs under varied viewing distance and
angle. Wu et al.\ \cite{wu2020cloak} and Xu et al.\ \cite{xu2020tshirt} extended
printed attacks to garments, the latter modelling cloth deformation explicitly
so that the perturbation survives non-rigid motion. In each case the printable
set is a fixed reference palette, and the reproduction step is treated as a
constraint to be satisfied rather than a quantity to be measured.

A parallel strand constrains the patch to look plausible rather than merely to
be printable. Duan et al.\ \cite{duan2020advcam} cast the constraint as style
transfer onto a natural texture, Hu et al.\ \cite{hu2021naturalistic} draw the
patch from the output manifold of a generative model, and Deng et al.\
\cite{deng2023ruststyle} restrict an aerial patch to a rust-like appearance so
that it reads as surface weathering. These constrain \emph{appearance}, which is
a perceptual criterion evaluated after the fact. The constraint studied here is
\emph{reproducibility}: whether a given colour can be laid down by a particular
printer on a particular substrate at all, which is a measurable property of
hardware and is imposed during optimisation rather than assessed afterwards. The
two are complementary, and \Cref{tab:levels} treats the appearance-side
constraints as the anchor and palette levels.

Physical vehicle camouflage generalises the planar patch to a full surface.
Wang et al.\ \cite{wang2021das} suppress model attention alongside human
attention, and Wang et al.\ \cite{wang2022fca} optimise a full-coverage texture
for multi-view robustness. Full-coverage camouflage is a different threat model
from the one considered here: a roof decal is a bounded, removable, flat region
that can be printed on adhesive vinyl, which is what makes the printer gamut
the binding constraint.

\subsection{Patch attacks in aerial and remote-sensing imagery}
Aerial imagery differs from ground-level imagery in viewpoint, object scale, and
the range of altitudes and viewing angles encountered; VisDrone
\cite{zhu2022visdrone} is a representative UAV-captured detection benchmark. Sun
et al.\ \cite{sun2023threatening} attacked object detection in optical
remote-sensing imagery, noting that such images contain many more objects per
frame than natural images, which makes a single patch less able to affect all of
them. Tang et al.\ studied patch attacks on aerial detectors under both
white-box \cite{tang2023aerial} and black-box \cite{tang2024blackbox} access,
the latter by differential evolution. Lian et al.\ \cite{lian2022benchmarking}
assembled a benchmark for patches against aerial detection and subsequently
showed \cite{lian2023cba} that a patch placed in the background rather than on
the target can suppress detection, which indicates that the roof-mounted threat
model adopted here is one of several. Beyond patches, Xu and Ghamisi
\cite{xu2022universal} constructed universal perturbations for remote-sensing
imagery, and Bai et al.\ \cite{bai2024stealthy} attacked semantic segmentation
in that domain, perturbing a chosen victim class while holding the predictions
for other classes within 2 per cent of their clean accuracy, which lowers the
rate at which the adversarial example is itself detected. That is the nearest
prior work on the segmentation side, a task the aerial patch literature has
addressed less often than detection.

It also marks the boundary of the present threat model. Both that work and
Xu and Ghamisi's operate on image pixels, and therefore assume an attacker who
can modify the image after capture. A printed decal modifies the scene instead,
so it must survive rendering, printing, illumination and resampling before it
reaches the model at all. The two are not competing methods; the pixel-space
results indicate what is achievable when the imaging chain is not in the way,
and the constraint set of \Cref{sec:method-realism} is an attempt to quantify
what that chain costs.

\paragraph{Scale.} Two studies bear directly on the band-limit argument of
\Cref{sec:method-realism}. Lu et al.\ \cite{lu2021scaleadaptive} observe that a
patch of fixed size is mismatched to targets whose apparent size varies, so that
a patch small enough not to occlude a small aircraft is too small to affect a
large one, and scale the patch to the target in response. Zhang et al.\
\cite{zhang2022multiscale} address the same multi-scale problem for UAV imagery.
Both treat scale as a property of the target to be adapted to. \Cref{eq:texels}
instead treats it as a constraint on the patch: at a given ground sample
distance only structure coarser than the sampling limit survives to the victim,
regardless of how the patch is scaled, which is why the realism levels are
defined in texels per image pixel rather than in pixels.

\paragraph{Physical evaluation.} Du et al.\ \cite{du2022physical} fabricated and
installed patches on and near cars and evaluated them against an overhead
detector while accounting for atmospheric conditions and observer-target
distance, reporting that the attack does transfer to the physical setting.
Shrestha et al.\ \cite{shrestha2023uav} pursued robustness of a UAV-directed
patch across capture conditions. Lian et al.\ \cite{lian2025padetbench} argue
that physical dynamics and cross-domain transformation cannot be strictly
regulated outdoors, so that reported comparisons between physical attacks are
often unaligned, and propose simulation as the means of obtaining matched
conditions. That argument is the reason the primary evaluation here is
simulated, with exact viewing geometry, and the physical arm is reported as a
bound rather than as a competing measurement.

Den Hollander et al.\ \cite{denhollander2020camouflage} optimised patches
intended to conceal vehicles from an aerial detector, and report that
effectiveness falls as the patch is viewed from further away, which is the same
scale dependence that \Cref{eq:texels} makes explicit here.

Two studies bear directly on the physical arm reported here. Hartnett et al.\
\cite{hartnett2022empirical} printed patches and evaluated them against overhead
object detection models, concluding that implementing the attack under those
conditions is substantially harder than under previously considered conditions
and that the real-world threat may be lower than earlier work suggested. Woo and
Lee \cite{woo2026digital} studied digital-to-physical transfer specifically for
aerial vehicle detection and report a large gap between digital and physical
effectiveness. The physical result in \Cref{sec:res-phys} is consistent in
direction with both, which is relevant to its interpretation: a null in this
setting is not isolated.

\subsection{Transfer between victims}
A patch optimised against one detector does not necessarily transfer to another.
Huang et al.\ \cite{huang2023tsea} treat this as an ensembling problem and
improve transfer by self-ensembling over augmented copies of a single model.
This study does not attempt to improve transfer; it measures it, reporting every
victim in \Cref{tab:net} against its own size-matched control so that transfer
and the effect of applying any object of the same size are not conflated.

\subsection{Victim architectures, simulation and interpretability}
The victims evaluated here span convolutional and transformer families:
SegFormer \cite{xie2021segformer}, and UPerNet decoders over Swin
\cite{liu2021swin} and ConvNeXt \cite{liu2022convnet} backbones, with decoders
initialised from ADE20K \cite{zhou2017ade20k}. The open-vocabulary segmenter
CLIPSeg \cite{luddecke2022clipseg} specifies its class by text prompt, which
permits the same patch to be re-tested against a different class description
without retraining. AirSim \cite{shah2018airsim} provides the rendering and
sensor model used to generate the corpus. Grad-CAM \cite{selvaraju2017grad}
localises the evidence supporting the vehicle class, and ArUco markers
\cite{garrido2014aruco} are used to rectify the printed colour chart. Perceptual
distance metrics such as LPIPS \cite{zhang2018unreasonable} offer an alternative
to the colourimetric measures used here.

\subsection{Position of this work}
This study is orthogonal to improvements in patch optimisation. The optimiser is
held fixed and the feasible set is changed, so that the reported attack rates
correspond to decals that could be manufactured and applied. Three choices
distinguish it from the work above. The printable set is measured for the
specific printer and substrate rather than taken as a fixed reference palette,
and that measurement is imposed during optimisation rather than used to score
the result afterwards. Every rate is reported against a size-matched control, so
that the contribution of the optimised content is separated from the
contribution of applying any object of comparable size. And the victim roster
covers semantic segmentation as well as detection, which the aerial patch
literature has addressed less often. To our knowledge the printer gamut has not
previously been measured for the specific device used and then imposed as an
optimisation constraint, although we note that the literature surveyed here is
not exhaustive and the surveys cited above should be consulted for coverage.

\FloatBarrier
\section{Methodology}\label{sec:method}

\subsection{Threat model and composition}\label{sec:method-threat}
An operator prints a planar decal and applies it to a single flat upper surface
of a vehicle. The decal must fit within that surface without overhang; a
composite spanning a cab, an intervening gap and a cargo bed does not correspond
to an applicable object. The sensor is airborne and near-nadir. The attacker has
white-box access to one semantic segmentation model (SegFormer-B0
\cite{xie2021segformer}) and one dense detector (RetinaNet
\cite{lin2017focal}); all remaining
victims are transfer targets.

Let $x \in [0,1]^{3 \times H \times W}$ be a frame, $\delta$ the patch texture,
and $T_\theta$ the geometric warp placing the decal on the vehicle's roof plane
under frame geometry $\theta$, producing a soft mask $\alpha_\theta$. The
composited frame is
\begin{equation}\label{eq:composite}
  x'(\delta,\theta) \;=\; \bigl(1-\alpha_\theta\bigr) \odot x
  \;+\; \alpha_\theta \odot C\bigl(T_\theta(\Pi(\delta))\bigr),
\end{equation}
where $\Pi$ is the realism projection of \Cref{sec:method-realism} and $C$ the
print chain of \Cref{sec:method-chain}. Both $\Pi$ and $C$ are differentiable,
so the composite remains end-to-end differentiable in $\delta$.

\subsection{Optimisation objective}\label{sec:method-obj}
The patch is optimised under EOT \cite{athalye2018synthesizing} over the joint
distribution of frames and print-chain draws:
\begin{equation}\label{eq:objective}
  \delta^{\star}
  \;=\; \arg\min_{\delta}\;
  \Exp_{(x,\theta)\sim\mathcal{D},\, C\sim\mathcal{C}}
  \Bigl[\;
     \mathcal{L}_{\text{atk}}\bigl(x'(\delta,\theta)\bigr)
     \;+\; \textstyle\sum_{j} w_j\, \mathcal{L}_j(\delta)
  \Bigr],
\end{equation}
where $\mathcal{L}_{\text{atk}}$ is the task term below and the
$\mathcal{L}_j$ are the realism terms of \Cref{sec:method-realism}, with weights
$w_j$ given in \Cref{tab:weights}.

\begin{table}[htbp]
  \centering \small
  \caption{Realism term weights $w_j$ in \Cref{eq:objective}. These shape a
  feasible iterate; they do not create one, since the projection $\Pi$ already
  enforces the constraint set.}
  \label{tab:weights}
  \begin{tabular}{lcl}
    \toprule
    Term & $w_j$ & Role \\
    \midrule
    $\mathcal{L}_{\text{ink}}$ (\Cref{eq:inkloss})      & 0.10 & soft distance to nearest printable ink \\
    $\mathcal{L}_{\text{flat}}$ (\Cref{eq:atv})         & 0.05 & anchored anisotropic total variation \\
    $\mathcal{L}_{\text{anchor}}$ (\Cref{eq:anchorloss})& 0.20 & hinged distance to the reference design \\
    $\mathcal{L}_{\text{gamut}}$ (\Cref{eq:gamutloss})  & 0.05 & penalty for leaving the measured gamut \\
    $\mathcal{L}_{\text{nps}}$                          & 0.02 & reported for comparability only \\
    \bottomrule
  \end{tabular}
\end{table}

For segmentation, let $p_{\text{veh}}$ be the victim's vehicle posterior and $m$
the ground-truth vehicle mask. The attack term is hinged at a threshold $\tau$,
so that gradient is not spent on pixels already suppressed below $\tau$:
\begin{equation}\label{eq:segloss}
  \mathcal{L}_{\text{seg}}
  \;=\;
  \frac{\sum_{u} \relu\bigl(p_{\text{veh}}(u) - \tau\bigr)\, m(u)}
       {\max\bigl(\sum_{u} m(u),\, 1\bigr)} ,
\end{equation}
with $u$ indexing pixels. Because \Cref{eq:segloss} hinges the posterior rather
than its log-odds, its gradient with respect to the vehicle logit carries a
factor $p_{\text{veh}}(1-p_{\text{veh}})$ and is small on pixels the victim
segments confidently. \Cref{sec:res-objcheck} tests whether this, or the choice
of pixels the term covers, limits the attack, and finds that it does not.
For detection the analogous term is taken over the
per-location vehicle response $q_\ell$ at each feature-pyramid level $\ell$,
deliberately in the same form so that the two are commensurate and can be summed
without an additional scale factor:
\begin{equation}\label{eq:detloss}
  \mathcal{L}_{\text{det}}
  \;=\;
  \frac{1}{|\mathcal{P}|}\sum_{\ell \in \mathcal{P}}
  \frac{\sum_{u} \relu\bigl(q_\ell(u) - \tau\bigr)\, m(u)}
       {\max\bigl(\sum_{u} m(u),\, 1\bigr)} .
\end{equation}
The joint objective used for the patches compared in \Cref{sec:results} is
$\mathcal{L}_{\text{atk}} = \mathcal{L}_{\text{seg}} + \mathcal{L}_{\text{det}}$,
the sum of \Cref{eq:segloss,eq:detloss} with no additional weighting. Both are
normalised by the same mask area, so neither term dominates by construction when
the victim resolutions differ.

\subsection{Nested realism constraints}\label{sec:method-realism}
Five nested constraint sets are defined, each a subset of the preceding one
(\Cref{tab:levels}).

\begin{table}[htbp]
  \centering \small
  \caption{Nested realism constraint sets. Each level adds one constraint to the
  preceding level.}
  \label{tab:levels}
  \begin{tabular}{cl}
    \toprule
    Level & Constraint set \\
    \midrule
    0 & Unconstrained pixels \\
    1 & Band limit \\
    2 & Band limit $+$ anchored total variation \\
    3 & Band limit $+$ anchored total variation $+$ palette \\
    4 & Band limit $+$ anchored total variation $+$ palette $+$ anchor design \\
    \bottomrule
  \end{tabular}
\end{table}

\paragraph{Band limit.} The number of patch texels falling within one image pixel
at the operating ground sample distance $g$ is
\begin{equation}\label{eq:texels}
  \rho \;=\; \frac{R\, g}{s},
\end{equation}
for a patch of $R$ texels across a physical side of $s$ metres. Texture finer
than two texels per image pixel is averaged away by the sensor, so the
projection applies a Gaussian low-pass of standard deviation set by $\rho$
before compositing. At $R = 256$, $s = \SI{2.0}{\metre}$ and
$g = \SI{0.056}{\metre\per\pixel}$ this gives $\rho = 7.17$.

\paragraph{Anchored total variation.} A plain total-variation penalty removes the
reference design's own boundaries along with the noise. The penalty is therefore
weighted down wherever the anchor $a$ has an edge:
\begin{equation}\label{eq:atv}
  \mathcal{L}_{\text{flat}}
  \;=\;
  \Exp_{u}\Bigl[\;
    \lVert \nabla \delta(u) \rVert \,
    \exp\!\bigl(-\lVert \nabla a(u) \rVert / \kappa\bigr)
  \Bigr],
\end{equation}
with $\kappa = 0.10$. Regions between the design's boundaries are driven toward
constant colour while the boundaries themselves survive.

\paragraph{Palette.} Colour distances throughout are CIEDE2000
\cite{luo2001ciede2000}, computed following the implementation notes and
reference test data of Sharma et al.\ \cite{sharma2005ciede2000}, which the
implementation used here was checked against. Let $\{c_k\}_{k=1}^{K}$ be the ink
set. A hard nearest-ink distance has zero gradient to every ink but the winner,
so a soft minimum over CIEDE2000 distances is used:
\begin{equation}\label{eq:inkloss}
  \mathcal{L}_{\text{ink}}
  \;=\;
  \Exp_{u}\Bigl[
    -\beta \log \textstyle\sum_{k} \exp\bigl(-\dE(\delta(u), c_k)/\beta\bigr)
  \Bigr] \big/ \Delta_{\text{ref}},
\end{equation}
with $\beta = 0.25$, at which the two nearest inks receive gradient and the
remainder effectively do not. A complementary term penalises colour lying
outside the ink set beyond a tolerance $\Delta_{\text{ref}}$:
\begin{equation}\label{eq:gamutloss}
  \mathcal{L}_{\text{gamut}}
  \;=\;
  \Exp_{u}\Bigl[
    \relu\bigl(\min_{k} \dE(\delta(u), c_k) - \Delta_{\text{ref}}\bigr)
  \Bigr] \big/ \Delta_{\text{ref}} .
\end{equation}

\paragraph{Anchor.} The objective is not to reproduce the reference design,
which carries no attack, but to prevent the recolouring from leaving a palette
that could plausibly be printed. The term is hinged at a budget $\Delta_a$ and is
free inside it:
\begin{equation}\label{eq:anchorloss}
  \mathcal{L}_{\text{anchor}}
  \;=\;
  \Exp_{u}\bigl[\relu\bigl(\dE(\delta(u), a(u)) - \Delta_a\bigr)\bigr]
  \big/ \Delta_a .
\end{equation}
At level~4 the anchor's spot colours are added to the ink set; without this the
projection cannot reproduce its own anchor and the realism target lies outside
the feasible set. The non-printability score of \cite{sharif2016accessorize} is
retained as $\mathcal{L}_{\text{nps}}$ for comparability with prior work and is
computed in RGB as originally defined, rather than in a perceptually uniform
space, so that the reported value is the quantity other studies report.

\begin{figure}[htbp]
  \centering
  \includegraphics[width=\linewidth]{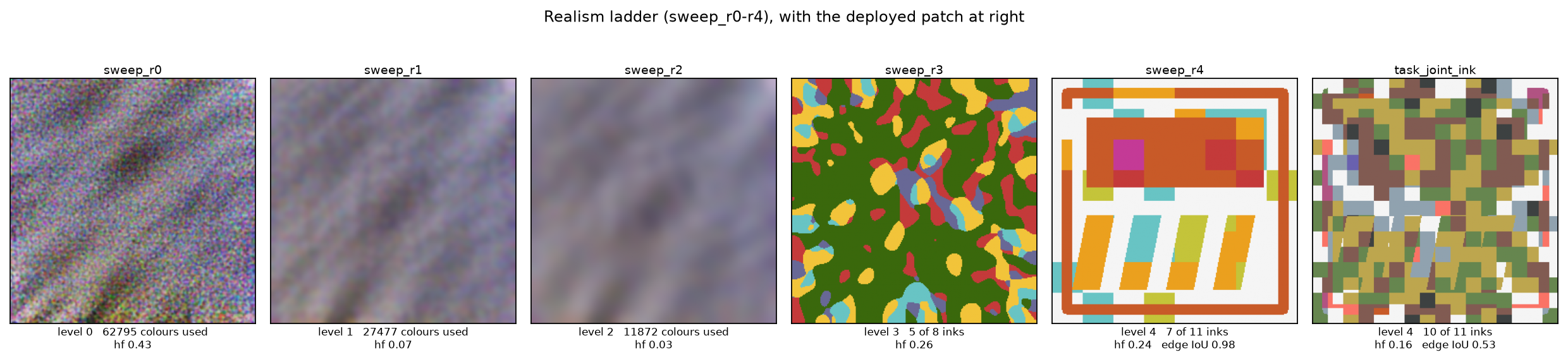}
  \caption{Patches produced at each realism level at matched patch size and
  optimisation budget, with the deployed patch at right. Annotations give the
  constraint level, the number of colours used out of the palette offered, the
  high-frequency energy ratio, and, where an anchor design exists, the fraction
  of its edges retained.}
  \label{fig:ladder}
\end{figure}

\Cref{fig:ladder} shows the resulting patches. The number of distinct colours in
a finished patch is reported as ``$k$ of $m$ inks'', where $m$ is the palette
size the manifold offered. Reporting $k$ alone would invite comparison against
the configured ink count, which would be misleading because the anchor's spot
colours legitimately enlarge the palette at level~4.

\subsection{Print-chain augmentation}\label{sec:method-chain}
\begin{figure}[htbp]
  \centering
  \includegraphics[width=\linewidth]{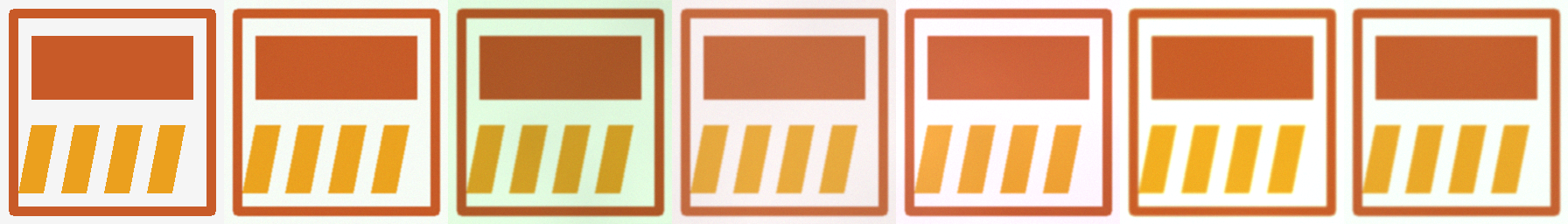}
  \caption{Seven independent draws from the print-chain augmentation applied to
  the reference cargo design. Each draw samples substrate reflectance, ink
  response and illumination.}
  \label{fig:printchain}
\end{figure}

The print chain $C$ in \Cref{eq:composite} models the transformations between a
digital texture and the pixels a model observes: quantisation to the ink set,
substrate reflectance, specular response and illumination variation. A draw is
sampled independently at each optimisation step, which is the
$C \sim \mathcal{C}$ expectation in \Cref{eq:objective}.
\Cref{fig:printchain} shows seven draws applied to the reference design.

\subsection{Measurement of the printer gamut}\label{sec:method-gamut}
The default ink set in comparable pipelines is a uniformly spaced RGB cube,
which does not correspond to the gamut of any physical printer. A printer
reproduces a bounded volume in colour space fixed by its inks and substrate,
and colours requested outside that volume are mapped onto its boundary by the
colour-management pipeline rather than reproduced \cite{morovic2001gamut}. The
mapping is specified by the device profile \cite{icc2010spec}, so the palette an
optimiser can actually realise is a property of the printer-substrate pair and
not of the colour space the optimiser works in. The chart procedure below
measures that pair directly, which avoids having to assume a rendering intent. A chart of the
216 requested colours was printed on the target printer and substrate, scanned,
and rectified by homography using four ArUco markers \cite{garrido2014aruco} at
its corners; the reproduced colour of each patch was then measured.

The measured set differs substantially from the cube: over the 216 chart
patches the median requested-to-reproduced distance is $\dE = 14.4$ with a 95th
percentile of $24.9$, and mean CIELAB chroma,
$C^{*} = \sqrt{(a^{*})^{2} + (b^{*})^{2}}$, falls from $60.1$ to $32.7$. The
reproducible palette is therefore approximately 46 per cent less chromatic than
the set the optimiser had been assuming.

\begin{figure}[htbp]
  \centering
  \includegraphics[width=0.78\linewidth]{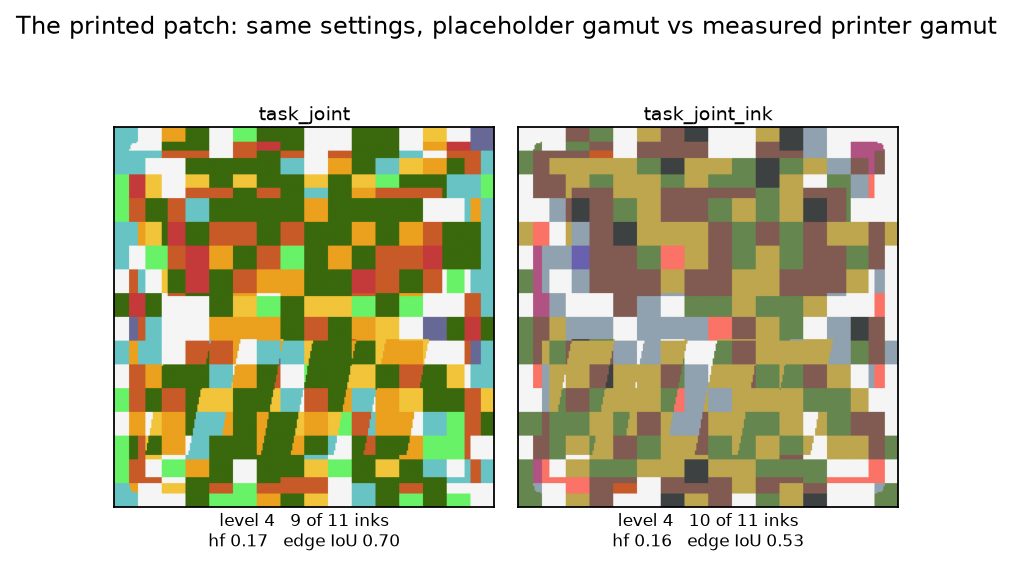}
  \caption{The controlled pair. Both patches share all seventeen recorded
  optimisation settings and differ only in the ink set: a placeholder colour
  cube (left) and the measured printer gamut (right). Median $\dE$ from patch
  colours to the nearest reproducible colour falls from $8.0$ to $0.0$.}
  \label{fig:printed}
\end{figure}

The ink set is an input to the optimisation and not solely a reporting artefact.
Measuring the gamut after optimisation does not alter the patch; it quantifies
the discrepancy between the palette assumed and the palette available.
\Cref{fig:printed} shows the two patches of the controlled pair. The median
$\dE$ from patch colours to the nearest reproducible colour falls from $8.0$
under the cube ink set to $0.0$ under the measured one, confirming that the
palette constraint binds: every colour the measured-gamut patch uses is one the
printer can lay down.

\subsection{Palette construction for the H1 experiments}\label{sec:method-palettes}
Testing whether chromatic extent or cardinality drives the result of
\Cref{sec:res-gamut} requires ink sets that vary one while holding the other.
Three families are built, all written as explicit sRGB lists so the reduction
inside the optimiser never fires and the cardinality is exact.

Let $\{c_k\}$ be the eight inks the measured chart reduces to, and write each
in CIELAB as $(L^{*}, C^{*}, h)$. The \emph{chroma} family holds $L^{*}$, $h$
and the cardinality fixed and scales $C^{*}$ by $\gamma \in \{0.5, 1, 1.5, 2\}$,
converting back to sRGB and clipping. Clipping means the achieved mean chroma
is not $\gamma$ times the original, so the achieved value is what
\Cref{tab:h1} reports: $16.4$, $32.8$, $47.7$ and $58.6$ against the cube ink
set's $54.6$. The \emph{cardinality} family draws $k \in \{3, 5, 8, 16, 32\}$
real colours from the same measured chart by $k$-means in CIELAB, so chroma
stays near the measured level by construction. The \emph{random} family keeps
the measured set's cardinality and its $L^{*}$ and $C^{*}$ values exactly and
redraws each hue angle uniformly, giving palettes matched on every summary
statistic the hypothesis appeals to and differing only in where the colours
sit.

\subsection{Physical capture and fiducial-free altitude recovery}\label{sec:method-phys}
Scaled physical capture is admissible because a photograph does not encode
absolute size. For an object of extent $L$ at range $d$ imaged with focal length
$f$ in pixels, the projected extent is
\begin{equation}\label{eq:scale}
  L_{\text{px}} \;=\; \frac{f\,L}{d},
\end{equation}
so reducing the object by a factor $s$ and the range by the same factor leaves
$L_{\text{px}}$, and hence the angular subtense and ground sample distance,
unchanged. Decals were printed at scale, applied to die-cast vehicles, and
photographed as clean/patched pairs from a single pose, so that illumination,
camera pose and automatic exposure are shared within a pair.

Recovery of camera pose from the printed fiducial frame was not possible in the
reported session: the frame was reassembled incorrectly, and its measured
width-to-height ratio was $0.55$ where the geometry requires $1.00$, with the
same ratio at every capture distance. Distance was therefore recovered by
inverting \Cref{eq:scale} using the known body length of the photographed
vehicle and the EXIF focal length,
\begin{equation}\label{eq:altrecover}
  \hat{d} \;=\; \frac{f\,L_{\text{veh}}}{L_{\text{px}}},
  \qquad
  \hat{h} \;=\; h_{\text{ref}}\,\frac{\hat{d}}{d_{\text{ref}}},
\end{equation}
where $\hat{h}$ is the effective altitude, and $h_{\text{ref}}$, $d_{\text{ref}}$
are the reference altitude and its scaled capture distance. No fiducial is
involved. The procedure generalises to any physical capture of an object of
known dimension and may be useful as a redundant channel even where fiducial
markers are available.

\subsection{Evaluation metrics}\label{sec:method-metrics}
An instance is termed \emph{attackable} if the victim locates it in the clean
frame: intersection over union (IoU) $\geq 0.50$ for segmentation, or a detection
with confidence $\geq 0.50$ matched at IoU $\geq 0.50$ for detection. Attack
success rate is computed over the attackable subset $\mathcal{A}$ only,
\begin{equation}\label{eq:asr}
  \asr \;=\; \frac{1}{|\mathcal{A}|}\sum_{i \in \mathcal{A}}
  \mathbb{1}\bigl[\text{victim fails on instance } i \text{ under attack}\bigr],
\end{equation}
and is reported as undefined, never as zero, where $\mathcal{A} = \emptyset$,
since a victim that did not locate the vehicle in the clean frame cannot be said
to have been attacked.

Each attack is paired with a size-matched control: the unoptimised reference
design, composited at the same metric size onto the same instances. Net attack
success is
\begin{equation}\label{eq:net}
  \net \;=\; \asr_{\text{attack}} - \asr_{\text{control}} .
\end{equation}

Coverage fraction, which governs how much of a panel the decal occupies, is
\begin{equation}\label{eq:coverage}
  \phi \;=\; \frac{s^{2}}{A_{\text{panel}}} .
\end{equation}

Confidence intervals are obtained by bootstrap over vehicles rather than over
instances, because repeated views of one parked vehicle are not independent
observations. For the physical evaluation, which returned zero successes, we
report the exact one-sided Clopper--Pearson upper bound at level $1-\alpha$ for
$0$ successes in $n$ trials,
\begin{equation}\label{eq:cp}
  p_{\max} \;=\; 1 - \alpha^{1/n}.
\end{equation}
The size of the held-out fleet is the binding constraint on every interval in
this paper, and it is a property of the corpus rather than of the analysis. The
corpus contains ten target vehicles, of which seven are used for optimisation
and three are held out, so the cluster bootstrap resamples three units. Three
is small enough to change the character of the interval: a resample draws the
same vehicle three times with probability $3 \times (1/3)^{3} = 1/9$, and those
degenerate draws set the lower tail. A 95 per cent vehicle-level interval can
therefore exclude zero only when the effect holds on all three vehicles
separately, which makes every interval reported here conservative and makes a
failure to exclude zero weak evidence of absence.

This is stated at the point the intervals are defined, rather than only in
\Cref{sec:limits}, because it bounds what the design can show before any result
is read. Enlarging the fleet is a rendering cost, not a methodological change:
the ten vehicle models are the assets available in the simulated city, and a
larger held-out set would require additional models and a re-render of the
corpus. \Cref{sec:limits} quantifies the consequences for specific claims.

\FloatBarrier
\section{Experimental Setup}\label{sec:setup}

\subsection{Dataset}\label{sec:setup-data}
Experiments use GeoPatchCity, a synthetic aerial corpus rendered in Unreal
Engine 4.27.2 under AirSim 1.8.1 \cite{shah2018airsim}, operated in
ComputerVision mode so that the camera is posed directly rather than flown under
a physics model. Each frame is therefore rendered at an exactly commanded
position and orientation, and the pose written to the annotation record is the
pose the renderer used rather than an estimate recovered from it. The
consequence relevant here is that viewing geometry is ground truth: the altitude
and nadir angle against which \Cref{sec:res-envelope} stratifies attack success
are exact, not inferred. The platform does not move, so motion blur is absent
from the renders. Illumination was varied over three fixed times of day
(morning, midday, evening) and weather held at a single clear condition.

\paragraph{Camera model.} The virtual camera is configured to an \SI{84}{\degree}
horizontal field of view (\SI{53.7}{\degree} vertical) at $1280 \times 720$
pixels, giving $f_x = f_y = \SI{710.79}{\pixel}$ with the principal point at
$(640.0, 360.0)$. Under the declared $6.17 \times \SI{3.47}{\milli\metre}$ sensor
this corresponds to a \SI{4.82}{\micro\metre} pixel pitch and a
\SI{3.43}{\milli\metre} focal length, or approximately \SI{20.9}{\milli\metre}
in 35\,mm-equivalent terms. This is within \SI{5}{\percent} of the
\SI{20}{\milli\metre} published for the DJI Phantom~4 \cite{dji2016phantom4},
one of the several Mavic- and Phantom-series platforms used to acquire VisDrone
\cite{zhu2022visdrone}; VisDrone was captured with a range of platforms rather
than a single one, so this is a representative rather than an exact match.

Intrinsics were verified against the renderer by reprojecting a $3 \times 3$
cluster of \SI{3.11}{\metre} square fiducial markers at \SI{10}{\metre} spacing,
vertically staggered over a \SI{3}{\metre} range. All nine markers were
segmented and observed from 16 verification viewpoints spanning $0$ to
\SI{70}{\degree} nadir angle at altitudes from $8$ to \SI{120}{\metre}, a range
exceeding the capture grid, yielding 115 marker observations. Verification
recovered a constant translation of $(-0.027, 0.014, 39.978)$\,m between the
frame in which the simulator reports object poses and the frame in which it
places the camera. Applying this correction reduced mean reprojection error by
three orders of magnitude, to \SI{0.873}{\pixel}. Solving jointly for focal
length under the corrected frame returned \SI{710.50}{\pixel}, a deviation of
\SI{0.041}{\percent} from the analytic value, so the analytic intrinsics were
retained.

\paragraph{Viewpoint sampling.} Viewpoints are sampled on a three-axis spherical
grid about each target: nine altitudes from $20$ to \SI{60}{\metre} in
\SI{5}{\metre} steps, twelve nadir angles from $0$ to \SI{70}{\degree}, and eight
azimuths at \SI{45}{\degree} intervals. The nadir axis is sampled at
\SI{10}{\degree} intervals to \SI{30}{\degree} and at \SI{5}{\degree} intervals
from $30$ to \SI{70}{\degree}, concentrating resolution where foreshortening
changes fastest. Commanded positions are jittered by up to $\pm\SI{2}{\metre}$ in
altitude and $\pm\SI{2}{\degree}$ in nadir angle, so no two frames share an
exactly repeated geometry; the record carries both the nominal grid cell and the
realised pose, and all stratification here is on the nominal cell while all
projection uses the realised pose. All 108 altitude-by-angle cells are
populated.

\paragraph{Targets and annotation.} Ten vehicles serve as targets, spanning the
silhouette and roof-area range of civilian traffic: two hatchbacks, two sports
cars, two SUVs, two pickups, a box truck and a truck cab. Each frame designates
one vehicle as the primary target and centres the viewpoint grid on it; other
vehicles in the field of view are annotated as secondary instances. Every frame
carries a per-instance vehicle mask, a nine-class semantic mask, and a geometric
record giving the intrinsic matrix, world-to-camera extrinsics, realised
altitude, nadir angle and azimuth, and the 3D bounding box of each annotated
vehicle in world and image coordinates, with per-face normals, per-frame ground
sample distance and a foreshortening factor.

Instances are filtered rather than silently dropped. A target below
\SI{150}{\pixel} visible area, or below a $0.25$ ratio of visible to projected
amodal area, is retained but marked \emph{ignored}; one below \SI{15}{\pixel} is
not recorded. \Cref{tab:dataset} summarises the resulting corpus.

\begin{table}[htbp]
  \centering \small
  \caption{GeoPatchCity summary and the subset evaluated in this study. Scale
  statistics are over primary targets, for which the viewpoint grid is centred.
  Split sizes are read from the distributed split files.}
  \label{tab:dataset}
  \begin{tabular}{llr}
    \toprule
    Group & Quantity & Value \\
    \midrule
    \multirow{5}{*}{Corpus}
      & Frames                              & \num{12562} \\
      & Annotated vehicle instances         & \num{21767} \\
      & \quad primary / secondary           & \num{12562} / \num{9205} \\
      & \quad class \emph{car} / \emph{truck} & \num{14662} / \num{7105} \\
      & Target vehicles                     & 10 \\
    \midrule
    \multirow{3}{*}{Labelling}
      & Fully labelled / ignored            & \num{18315} / \num{3452} \\
      & Truncated by image boundary         & \num{890} \\
      & Median visibility ratio / mask-box IoU & 0.416 / 0.625 \\
    \midrule
    \multirow{4}{*}{Scale}
      & Ground sample distance (\si{\metre\per\pixel}) & 0.028--0.254, median 0.075 \\
      & Slant range (\si{\metre})           & 20.0--180.5 \\
      & Median visible area, \SIrange{20}{60}{\metre} (\si{\pixel}) & \num{5455} $\rightarrow$ \num{694} \\
      & Foreshortening, \SIrange{0}{70}{\degree} & 1.00 $\rightarrow$ 0.34 \\
    \midrule
    \multirow{2}{*}{Splits}
      & Train frames / instances            & \num{9039} / \num{16644} \\
      & Holdout frames / instances          & \num{3523} / \num{5123} \\
    \midrule
    \multirow{2}{*}{This study}
      & Optimisation instances (near-nadir) & \num{855} \\
      & Evaluation instances                & \num{2534} \\
    \bottomrule
  \end{tabular}
\end{table}

\paragraph{Subset evaluated here.} This study does not use every annotated
instance. A patch must be placed on a roof panel that is both visible and large
enough to carry it, so the index applied to each split retains only primary
targets with at least \SI{900}{\pixel} of visible area and at least
\SI{250}{\pixel} of projected roof quadrilateral. Optimisation additionally
restricts to near-nadir frames, defined as nadir angle $\leq \SI{8}{\degree}$,
since that is the viewpoint the threat model assumes. These gates yield
\num{855} optimisation instances and a \num{2534}-instance evaluation set
spanning $0$ to \SI{70}{\degree} of nadir angle and $19$ to \SI{63}{\metre} of
altitude over three held-out vehicles (box truck, $n = \num{1199}$; two pickups,
$n = \num{698}$ and $n = \num{637}$). The altitude range exceeds the nominal
\SIrange{20}{60}{\metre} grid because of the $\pm\SI{2}{\metre}$ commanded
jitter. All attack rates in \Cref{sec:results} are computed on this evaluation
set.

%

\subsection{Victims and patches}\label{sec:setup-victims}
\paragraph{Victims.} Eleven models are evaluated, chosen to span convolutional
and transformer families and both dense and two-stage detection. CLIPSeg is
evaluated under three prompts, giving thirteen victim conditions in total;
\asr{} is reported per condition throughout. \Cref{tab:victims} lists them with
their primary references.

Six are closed-set semantic segmentation models: FCN-R50 \cite{long2015fcn},
DeepLabv3-R101 \cite{chen2017deeplabv3}, UPerNet \cite{xiao2018upernet}
decoders over ConvNeXt-T \cite{liu2022convnet} and Swin-T \cite{liu2021swin}
backbones, and SegFormer-B0 and SegFormer-B2 \cite{xie2021segformer}. The two
ResNet-backed models use the backbone of He et al.\ \cite{he2016resnet}. One is
an open-vocabulary segmenter, CLIPSeg \cite{luddecke2022clipseg}, evaluated
under three prompts. Four are detectors: RetinaNet \cite{lin2017focal},
Faster~R-CNN \cite{ren2017faster}, FCOS \cite{tian2019fcos} and YOLOv8n
\cite{jocher2023ultralytics,hussain2024yolo}. The three torchvision detectors
use feature pyramid networks \cite{lin2017fpn}. YOLOv8 was released as software
without an originating publication, so it is cited twice: the software record
\cite{jocher2023ultralytics} identifies the implementation actually run
(version 8.4.150), and Hussain \cite{hussain2024yolo} provides the
peer-reviewed account of its architecture. Redmon et al.\
\cite{redmon2016yolo} is cited for the origin of the family, not as a
description of the version used.

Segmentation decoders and detector heads were fine-tuned on the corpus from
ADE20K \cite{zhou2017ade20k} and COCO initialisations respectively. CLIPSeg is
not fine-tuned, since its class specification is a text prompt.

\begin{table}[htbp]
  \centering \small
  \caption{The victim roster: eleven models, giving thirteen victim conditions
  once the three CLIPSeg prompts are counted separately. White-box access is
  used for SegFormer-B0 and RetinaNet during optimisation of the compared
  patches; all other victims are evaluated by transfer only. The two UPerNet
  models are additionally attacked white box in \Cref{sec:res-objcheck}.}
  \label{tab:victims}
  \begin{tabular}{lllc}
    \toprule
    Victim & Family & Reference & Access \\
    \midrule
    \multicolumn{4}{l}{\emph{Closed-set semantic segmentation}} \\
    FCN-R50            & CNN         & \cite{long2015fcn,he2016resnet}   & transfer \\
    DeepLabv3-R101     & CNN         & \cite{chen2017deeplabv3,he2016resnet} & transfer \\
    UPerNet-ConvNeXt-T & CNN         & \cite{xiao2018upernet,liu2022convnet} & transfer \\
    UPerNet-Swin-T     & transformer & \cite{xiao2018upernet,liu2021swin}    & transfer \\
    SegFormer-B0       & transformer & \cite{xie2021segformer}           & white box \\
    SegFormer-B2       & transformer & \cite{xie2021segformer}           & transfer \\
    \midrule
    \multicolumn{4}{l}{\emph{Open-vocabulary segmentation}} \\
    CLIPSeg ($\times 3$ prompts) & transformer & \cite{luddecke2022clipseg} & transfer \\
    \midrule
    \multicolumn{4}{l}{\emph{Object detection}} \\
    RetinaNet   & dense, CNN     & \cite{lin2017focal,lin2017fpn} & white box \\
    FCOS        & dense, CNN     & \cite{tian2019fcos,lin2017fpn} & transfer \\
    Faster R-CNN & two-stage, CNN & \cite{ren2017faster,lin2017fpn} & transfer \\
    YOLOv8n     & dense, CNN     & \cite{jocher2023ultralytics,hussain2024yolo} & transfer \\
    \bottomrule
  \end{tabular}
\end{table}

\paragraph{The controlled pair.} Two patches are compared throughout.
\Cref{tab:settings} lists every recorded optimisation setting for both. They
agree on all seventeen and differ only in the ink set. The axes are stated
explicitly rather than summarised as ``matched'', because an earlier revision of
this work compared two patches that in fact differed in task, optimisation
budget, batch size and band design size; \Cref{tab:net} replaces that
comparison.

\begin{table}[htbp]
  \centering \small
  \caption{Every recorded optimisation setting for the controlled pair. The two
  runs agree on all seventeen axes and differ only in the ink set (final row).}
  \label{tab:settings}
  \begin{tabular}{llcc}
    \toprule
    Group & Setting & Cube run & Measured run \\
    \midrule
    \multirow{6}{*}{Objective}
      & Task                   & joint & joint \\
      & Segmentation victim    & SegFormer-B0 & SegFormer-B0 \\
      & Detection victim       & RetinaNet & RetinaNet \\
      & Realism level          & 4 & 4 \\
      & Anchor design          & cargo & cargo \\
      & Print chain            & enabled & enabled \\
    \midrule
    \multirow{5}{*}{Geometry}
      & Patch size             & \SI{2.0}{\metre} & \SI{2.0}{\metre} \\
      & Size mode              & fixed & fixed \\
      & Texture resolution     & $256^{2}$ & $256^{2}$ \\
      & Band design size       & \SI{2.0}{\metre} & \SI{2.0}{\metre} \\
      & Ground sample distance & \num{0.056} & \num{0.056} \\
    \midrule
    \multirow{5}{*}{Optimisation}
      & Steps                  & \num{3000} & \num{3000} \\
      & Batch size             & 1 & 1 \\
      & Learning rate          & \num{0.02} & \num{0.02} \\
      & Anchor supercell       & 16 & 16 \\
      & Training instances     & \num{855} & \num{855} \\
    \midrule
    Capacity & Degrees of freedom & \num{1386} & \num{1386} \\
    \midrule
    \textbf{Varied} & \textbf{Ink set} & \textbf{colour cube}
                    & \textbf{measured gamut} \\
    \bottomrule
  \end{tabular}

  \vspace{2pt}
  {\footnotesize Ground sample distance in \si{\metre\per\pixel}. Degrees of
  freedom is the number of values the optimiser controls after the level-4
  projection.\par}
\end{table}

\paragraph{Seed replication.} Because the two patches are single optimisation
runs, the comparison was repeated at two further random seeds per ink set, with
all other settings fixed, giving three runs per condition
(\Cref{sec:res-seeds}).

\paragraph{Reproducibility.} All results are computed on a single
2534-instance holdout. Earlier revisions of this work evaluated different runs at
strides 4, 6 and 8, yielding four non-nested subsets whose attack rates were not
mutually comparable; every run has been re-scored at stride~1. The one
exception is the diagnostic comparison of attack terms in \Cref{tab:objvariants},
which is scored on a single stride-5 subset shared by all of its rows and is
labelled as such. Configuration, seed, loss history and per-run realism metrics
are archived with the results.

\FloatBarrier
\section{Results}\label{sec:results}

\subsection{Effect of the measured gamut}\label{sec:res-gamut}
\Cref{tab:net} reports $\net$ (\Cref{eq:net}) for the controlled pair against
every victim admitting a control, from one optimisation run per condition. On
all six closed-set segmentation victims the patch optimised under the measured
gamut achieved the higher net attack success, despite drawing on a palette
approximately 46 per cent less chromatic. The ordering is reversed on the
open-vocabulary segmenter, by $+0.129$ and $+0.124$ under the two prompts with
a defined attackable subset, and on all four detectors, by $0.002$ to $0.021$.

\Cref{tab:net} is reported as the single-run result it is, and
\Cref{sec:res-seeds} replicates every row of it over three seeds. That
replication should be read before any row here is treated as an ordering. It
sustains three of the six closed-set segmentation rows, leaves two unresolved,
and reverses the sign on the sixth; it confirms both open-vocabulary rows; and
it shows the detection rows to be better resolved than the single run
suggested, with the cube patch the stronger of the two on two of four
detectors. The rows of \Cref{tab:net} that survive replication are marked
there.

\begin{table}[htbp]
  \centering \small
  \caption{Net attack success rate, $\net = \asr - \asr_{\text{control}}$
  (\Cref{eq:net}), on a single 2534-instance holdout. Both patches share all
  seventeen recorded optimisation settings (\Cref{tab:settings}) and differ only
  in the ink set. Each is scored against its own size-matched control. Bold
  marks the higher net per victim in this single run. A dagger marks a victim
  whose three-seed ranges overlap in \Cref{tab:seeds}, where every victim in
  this table was replicated; those rows are not resolved and should not be read
  as an ordering, whichever way the single run fell. The daggers are therefore
  measured rather than extrapolated. Note that on UPerNet-Swin-T the sign
  reverses under replication, so the bold entry in that row records this run
  and not a claim. The CLIPSeg prompt ``a vehicle seen from above'' is
  omitted because its attackable subset is empty under both patches, making
  \asr{} undefined by \Cref{eq:asr}.}
  \label{tab:net}
  \begin{tabular}{llrrr rrr}
    \toprule
    & & \multicolumn{3}{c}{Cube ink set}
      & \multicolumn{3}{c}{Measured ink set} \\
    \cmidrule(lr){3-5}\cmidrule(lr){6-8}
    Arm & Victim & \asr{} & ctrl & $\net$ & \asr{} & ctrl & $\net$ \\
    \midrule
    \multirow{8}{*}{Seg.}
      & DeepLabv3-R101      & 0.667 & 0.173 & $+0.494$ & 0.768 & 0.173 & $\mathbf{+0.596}$ \\
      & FCN-R50             & 0.187 & 0.012 & $+0.175$ & 0.193 & 0.012 & $\mathbf{+0.181}^{\dagger}$ \\
      & SegFormer-B0        & 0.101 & 0.001 & $+0.100$ & 0.127 & 0.001 & $\mathbf{+0.127}$ \\
      & UPerNet-ConvNeXt-T  & 0.046 & 0.010 & $+0.036$ & 0.058 & 0.010 & $\mathbf{+0.048}^{\dagger}$ \\
      & UPerNet-Swin-T      & 0.037 & 0.003 & $+0.034$ & 0.039 & 0.003 & $\mathbf{+0.036}^{\dagger}$ \\
      & SegFormer-B2        & 0.013 & 0.000 & $+0.013$ & 0.017 & 0.000 & $\mathbf{+0.017}$ \\
      & CLIPSeg, ``a car''  & 0.562 & 0.326 & $\mathbf{+0.235}$ & 0.433 & 0.326 & $+0.106$ \\
      & CLIPSeg, ``a parked car'' & 0.545 & 0.370 & $\mathbf{+0.175}$ & 0.421 & 0.370 & $+0.051$ \\
    \midrule
    \multirow{4}{*}{Det.}
      & RetinaNet (white box) & 0.053 & 0.034 & $+0.019^{\dagger}$ & 0.051 & 0.034 & $+0.017^{\dagger}$ \\
      & Faster R-CNN          & 0.032 & 0.011 & $+0.021^{\dagger}$ & 0.029 & 0.011 & $+0.018^{\dagger}$ \\
      & FCOS                  & 0.068 & 0.049 & $\mathbf{+0.019}$ & 0.053 & 0.049 & $+0.004$ \\
      & YOLOv8n               & 0.063 & 0.037 & $\mathbf{+0.026}$ & 0.041 & 0.037 & $+0.005$ \\
    \bottomrule
  \end{tabular}
\end{table}

\begin{figure}[htbp]
    \centering
    \includegraphics[
        width=\textwidth,
        height=0.85\textheight,
        keepaspectratio
    ]{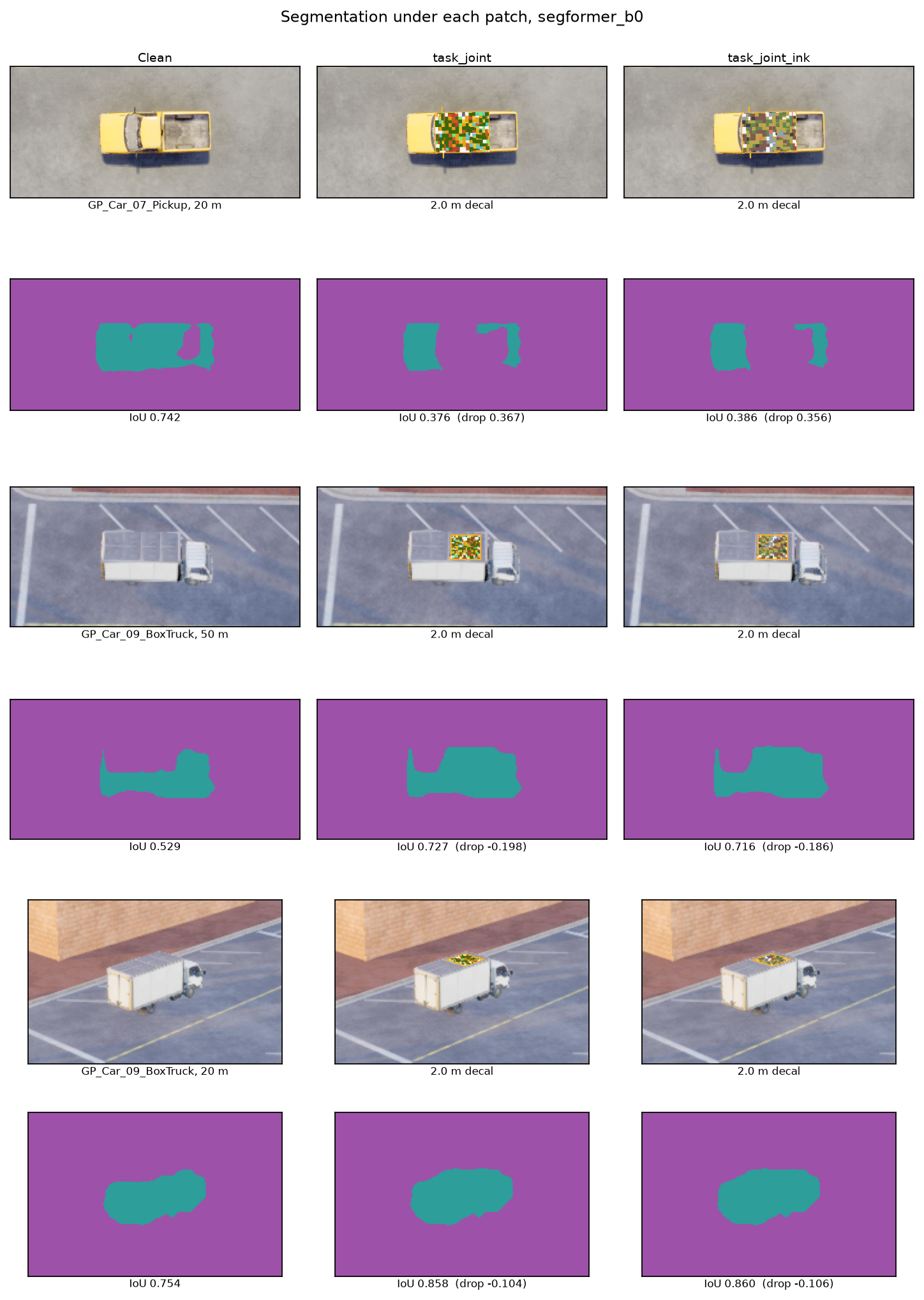}
    \caption{Semantic segmentation under each ink set, SegFormer-B0. Columns
    correspond to the same instance under the clean frame, the cube-constrained
    patch, and the measured-gamut patch; rows alternate input image and predicted
    vehicle mask. IoU and its change relative to the clean frame are given beneath
    each prediction.}
    \label{fig:compare-seg}
\end{figure}

\Cref{fig:compare-seg} shows the qualitative effect on matched instances. Both
patches reduce IoU on the pickup. On the box truck, where coverage
(\Cref{eq:coverage}) is approximately $0.18$, both increase IoU relative to the
clean frame: the predicted vehicle region grows. This is a property of the
victim on that vehicle rather than of the attack. The clean prediction misses
much of the featureless box body (mean clean IoU $0.59$ over its 474 attackable
instances), and any object on the roof supplies vehicle evidence there: a grey
square raises IoU on 96 per cent of box-truck instances and the unoptimised
decal on 98 per cent, by a mean of $0.085$ and $0.153$. The measured-gamut
patch raises it by $0.128$, less than the unoptimised decal, so the optimisation
acts against the effect but does not overcome it.

\subsection{Seed replication}\label{sec:res-seeds}
\Cref{tab:net} rests on one optimisation run per condition. To separate the
ordering from run-to-run variation, both conditions were repeated at two
further seeds with all other settings fixed, and all three runs per condition
were scored against every victim. \Cref{tab:seeds} gives the result for the
full roster. The size-matched controls do not depend on the seed and are those
of \Cref{tab:net}, so differences in \asr{} are differences in $\net$.

Replication does not support a uniform ordering, and the single-run result of
\Cref{sec:res-gamut} should be read against it. Of the six closed-set
segmentation victims, three separate cleanly: DeepLabv3-R101 at $+0.120$,
SegFormer-B0 at $+0.041$ and SegFormer-B2 at $+0.005$, in each case with the
three-seed ranges of the two conditions disjoint. Two more, FCN-R50 at $+0.025$
and UPerNet-ConvNeXt-T at $+0.010$, favour the measured gamut in the mean but
have overlapping ranges, so the ordering is not resolved at three seeds. On
UPerNet-Swin-T the sign reverses, the cube mean exceeding the measured mean by
$0.003$, which is $0.7$ pooled standard deviations and well inside seed noise.
The defensible statement is therefore that the measured gamut raises net attack
success on three of six closed-set segmentation victims, is consistent with
doing so on two more, and is indistinguishable on the sixth---not that it does
so on all six.

The other two arms replicate cleanly and in the opposite direction. Both
CLIPSeg prompts separate at $-0.120$ and $-0.122$, confirming that the cube
patch is the stronger of the two against the open-vocabulary segmenter. On
detection, FCOS at $-0.010$ and YOLOv8n at $-0.018$ separate, RetinaNet at
$-0.002$ does not, and Faster R-CNN separates by a margin too small to rely on.
This revises the reading offered before replication: the two ink sets are not
merely indistinguishable on the detection arm, the cube patch is reliably the
stronger one on two of four detectors. Because every detector net is at most
$+0.026$ and the controls run from $0.011$ to $0.049$, what separates is an
ordering between two small quantities, and \Cref{sec:res-det} should be read
first.

Two features of the design bound all of this. Three seeds per condition give an
exact permutation test only ten distinct splits, so its smallest attainable
two-sided $p$-value is $0.10$; separation of the ranges is reported as the
criterion because no stronger one is available at this sample size. And the
seeds vary the optimisation only. Victim weights, the instance set and the
controls are fixed across all six runs, so \Cref{tab:seeds} bounds
optimisation variance, not variance over victims or data.

\begin{table}[htbp]
  \centering \small
  \caption{Seed replication across the full victim roster: three independent optimisation runs per ink set, all other settings fixed. Means are over the three seeds and the range is their minimum and maximum. $\Delta$ is the measured-gamut mean minus the cube mean, so a positive value favours the measured gamut. ``Sep.'' marks victims whose three-seed ranges do not overlap; where they overlap the ordering is not resolved at this number of seeds and is not claimed. Controls are seed-independent and are those of \Cref{tab:net}.}
  \label{tab:seeds}
  \begin{tabular}{llrcrcrr c}
    \toprule
    & & \multicolumn{2}{c}{Cube ink set} & \multicolumn{2}{c}{Measured ink set} & & & \\
    \cmidrule(lr){3-4}\cmidrule(lr){5-6}
    Arm & Victim & mean & range & mean & range & $\Delta$ & $|\Delta|/s$ & Sep. \\
    \midrule
    \multirow{6}{*}{Closed-set seg.}
      & DeepLabv3-R101             & 0.655 & 0.647--0.667 & 0.775 & 0.768--0.785 & $\mathbf{+0.120}$ & 12.6 & \checkmark \\
      & FCN-R50                    & 0.187 & 0.180--0.195 & 0.213 & 0.193--0.228 & ${+0.025}$ & 1.9 &  \\
      & SegFormer-B0               & 0.106 & 0.098--0.118 & 0.147 & 0.127--0.172 & $\mathbf{+0.041}$ & 2.3 & \checkmark \\
      & UPerNet-ConvNeXt-T         & 0.050 & 0.045--0.060 & 0.060 & 0.058--0.064 & ${+0.010}$ & 1.6 &  \\
      & UPerNet-Swin-T             & 0.037 & 0.034--0.042 & 0.035 & 0.030--0.039 & ${-0.003}$ & 0.7 &  \\
      & SegFormer-B2               & 0.012 & 0.009--0.013 & 0.017 & 0.014--0.018 & $\mathbf{+0.005}$ & 2.5 & \checkmark \\
    \midrule
    \multirow{2}{*}{Open-vocab.}
      & CLIPSeg, ``a car''         & 0.553 & 0.545--0.562 & 0.434 & 0.424--0.444 & $\mathbf{-0.120}$ & 13.1 & \checkmark \\
      & CLIPSeg, ``a parked car''  & 0.536 & 0.519--0.545 & 0.414 & 0.396--0.424 & $\mathbf{-0.122}$ & 8.2 & \checkmark \\
    \midrule
    \multirow{4}{*}{Detection}
      & RetinaNet (white box)      & 0.054 & 0.052--0.056 & 0.051 & 0.051--0.052 & ${-0.002}$ & 1.2 &  \\
      & Faster R-CNN               & 0.034 & 0.029--0.040 & 0.027 & 0.026--0.029 & $\mathbf{-0.007}$ & 1.7 & \checkmark$^{\dagger}$ \\
      & FCOS                       & 0.066 & 0.065--0.068 & 0.056 & 0.053--0.061 & $\mathbf{-0.010}$ & 3.1 & \checkmark \\
      & YOLOv8n                    & 0.062 & 0.062--0.063 & 0.044 & 0.041--0.049 & $\mathbf{-0.018}$ & 6.4 & \checkmark \\
    \bottomrule
  \end{tabular}

  \vspace{2pt}
  {\footnotesize $s$ is the pooled between-seed standard deviation of the two conditions. With three seeds per condition an exact two-sided permutation test admits only ten distinct splits, so its smallest attainable $p$-value is $0.10$; every row marked separated attains it, and no row can do better under this design. The ratio $|\Delta|/s$ is therefore reported alongside, as a descriptive effect size rather than a test statistic. $^{\dagger}$Faster R-CNN's ranges clear each other by only $0.0004$, which the three-decimal display hides; that separation is marginal and the row should be read as unresolved.\par}
\end{table}

\subsection{Testing the regularisation hypothesis}\label{sec:res-h1}
\begin{figure}[htbp]
  \centering
  \includegraphics[width=\linewidth]{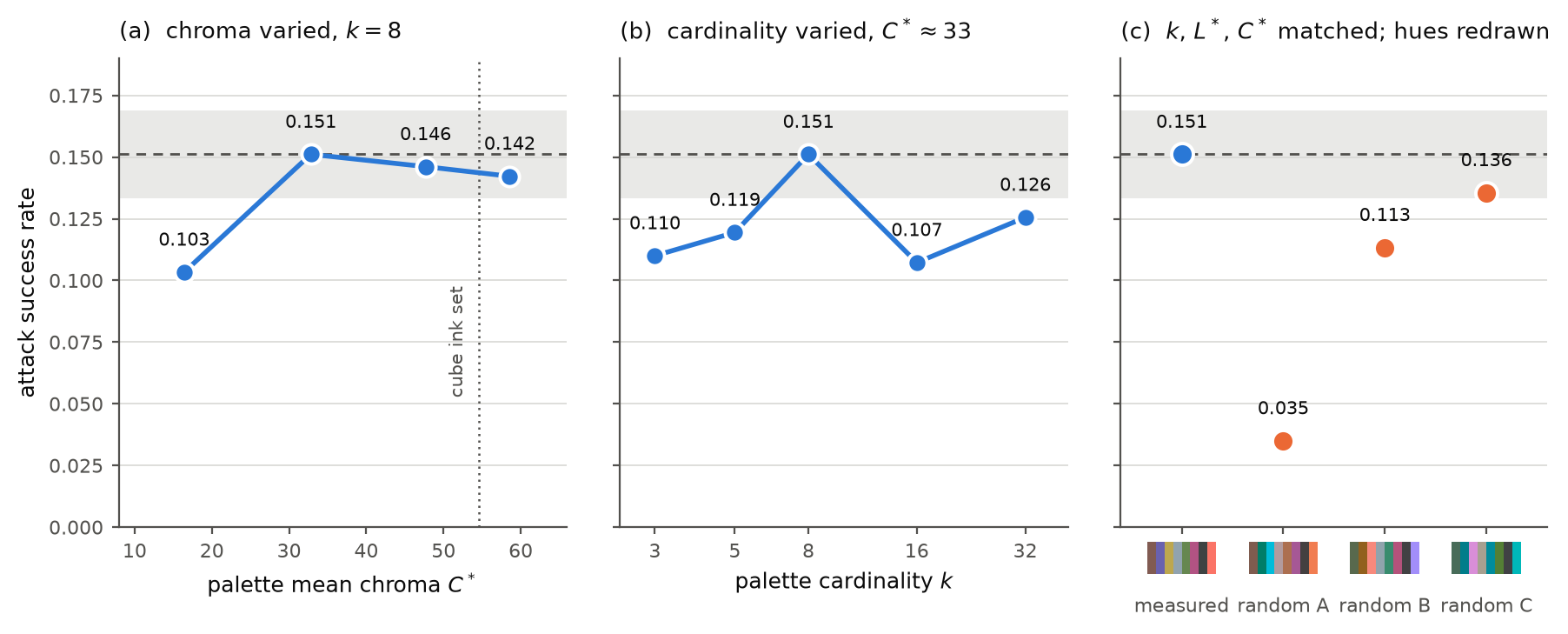}
  \caption{The palette ablation of \Cref{tab:h1}, on one axis. The grey band
  on each panel is the measured-gamut run plus or minus the between-seed
  standard deviation of $0.0178$ for this victim, so a point inside it is not
  separable from run-to-run variation. (a) Beyond the measured level, chroma
  does not matter: the two palettes bracketing the cube ink set's own mean
  chroma of $54.6$ sit inside the band. (b) Cardinality shows no trend across
  a tenfold range. (c) Three palettes matched to the measured set on
  cardinality, lightness and chroma, differing only in hue placement, span
  $0.035$ to $0.136$. Swatches beneath each point are the palettes themselves.}
  \label{fig:h1}
\end{figure}

\Cref{sec:res-gamut} raises an interpretation: that the measured gamut helps
because a palette a printer can reproduce is also a less chromatic and
lower-frequency one, and that \Cref{eq:texels} removes high-frequency structure
at altitude regardless. The controlled pair cannot test it, because the
measured set differs from the cube in both chromatic extent and cardinality and
either would reduce the optimiser's freedom. Eleven further patches were
optimised to separate them, each matching the measured-gamut run in all
seventeen recorded settings and differing only in the ink set
(\Cref{tab:h1,fig:h1}). Palettes are constructed by \Cref{sec:method-palettes}.

\paragraph{Chromatic extent does not account for the result.} Holding
cardinality at eight inks and scaling chroma about the neutral axis, attack
success is $0.151$ at the measured level of $C^{*} = 32.8$, $0.146$ at $47.7$
and $0.142$ at $58.6$. The last two bracket the cube ink set's own mean chroma
of $54.6$, and all three lie within half a between-seed standard deviation of
one another. A palette as chromatic as the cube therefore attacks as well as
the measured one once cardinality is matched, which the hypothesis predicts it
should not. Only the strongly desaturated palette at $C^{*} = 16.4$ is worse,
at $0.103$, so the relationship is not monotone in the direction proposed: too
little chroma costs attack success, and beyond the measured level additional
chroma neither helps nor hurts.

\paragraph{Cardinality does not account for it either.} Holding chroma near the
measured level and varying the number of inks from three to thirty-two gives
$0.110$, $0.119$, $0.151$, $0.107$ and $0.126$, a correlation with $k$ of
$+0.02$. Four of the five sit between $1.4$ and $2.5$ standard deviations below
the measured palette with no ordering in $k$, which is what an absence of
dependence looks like at this sample size.

\paragraph{Which colours, not how many or how saturated.} Three random palettes
matched to the measured set on cardinality, lightness and chroma score $0.035$,
$0.113$ and $0.136$. That spread has a standard deviation of $0.053$, three
times the between-seed variation of the optimiser itself, and it is the largest
effect anywhere in this experiment. Holding every summary statistic of the
palette fixed and moving only where its colours sit in the hue circle changes
attack success more than scaling chroma by a factor of four or changing
cardinality tenfold.

\paragraph{Conclusion.} H1 is not supported. The advantage of the measured
gamut in \Cref{tab:net} is not explained by its lower chromatic extent, which a
matched-cardinality palette at cube-level chroma reproduces without loss, nor
by its smaller cardinality, which shows no trend. What the evidence points to
instead is that the particular colours a printer happens to reproduce matter,
and that the summary statistics conventionally used to describe a palette do
not capture what the optimiser exploits. Why the measured set should be a
favourable set of colours is not established here, and the mechanism is
recorded as open rather than replaced with a second untested account.

Three limitations bound this. Each palette was optimised once, so differences
below roughly $0.04$ are not separable from seed noise, and only the
random-palette spread and the desaturated palette clear that bar decisively.
The random-palette result itself implies that between-palette variance exceeds
between-seed variance, so single runs per palette understate the uncertainty on
every row of \Cref{tab:h1}. And only SegFormer-B0 was tested; whether the same
holds for the victims where the measured gamut did not separate in
\Cref{tab:seeds} is unknown.

\begin{table}[htbp]
  \centering \small
  \caption{Testing hypothesis H1 on SegFormer-B0. Every run matches the measured-gamut patch of \Cref{tab:settings} in all seventeen recorded settings except the ink set. $C^{*}$ is the palette's mean CIELAB chroma after conversion to sRGB, $k$ its cardinality. The measured gamut appears in the first two blocks as the shared reference point. $\delta$ is the difference from it, expressed in units of the between-seed standard deviation of $0.0178$ measured for this victim in \Cref{tab:seeds}; a value below about $2$ is not separable from run-to-run variation.}
  \label{tab:h1}
  \begin{tabular}{llrrrrr}
    \toprule
    Experiment & Ink set & $k$ & $C^{*}$ & \asr{} & $\net$ & $\delta/s$ \\
    \midrule
    \multirow{4}{*}{\shortstack[l]{Chroma varied,\\$k = 8$ throughout}}
      & half chroma            &  8 &  16.4 & 0.103 & 0.102 & -2.7 \\
      & measured gamut$^{*}$   &  8 &  32.8 & 0.151 & 0.150 & --- \\
      & 1.5$\times$ chroma     &  8 &  47.7 & 0.146 & 0.145 & -0.3 \\
      & 2$\times$ chroma       &  8 &  58.6 & 0.142 & 0.141 & -0.5 \\
    \midrule
    \multirow{5}{*}{\shortstack[l]{Cardinality varied,\\$C^{*} \approx 33$ throughout}}
      & 3 inks                 &  3 &  31.9 & 0.110 & 0.109 & -2.3 \\
      & 5 inks                 &  5 &  31.7 & 0.119 & 0.118 & -1.8 \\
      & measured gamut$^{*}$   &  8 &  32.8 & 0.151 & 0.150 & --- \\
      & 16 inks                & 16 &  34.0 & 0.107 & 0.106 & -2.5 \\
      & 32 inks                & 32 &  33.7 & 0.126 & 0.125 & -1.4 \\
    \midrule
    \multirow{3}{*}{\shortstack[l]{Random palettes,\\$k$, $L^{*}$, $C^{*}$ matched}}
      & random A               &  8 &  30.2 & 0.035 & 0.034 & -6.5 \\
      & random B               &  8 &  32.8 & 0.113 & 0.112 & -2.1 \\
      & random C               &  8 &  27.5 & 0.136 & 0.135 & -0.9 \\
    \bottomrule
  \end{tabular}

  \vspace{2pt}
  {\footnotesize $^{*}$The measured gamut is the reference for $\delta$. Run independently here, it scores $0.151$ against the $0.127$--$0.172$ three-seed range of the same condition in \Cref{tab:seeds}, so it reproduces that condition. The ink reduction inside the pipeline is an unseeded $k$-means over the chart, so the eight inks are not bit-identical between the two runs; all comparisons in this table are therefore made within it.\par}
\end{table}

\subsection{Realism constraints and attack success}\label{sec:res-realism}
\begin{figure}[htbp]
  \centering
  \includegraphics[width=0.72\linewidth]{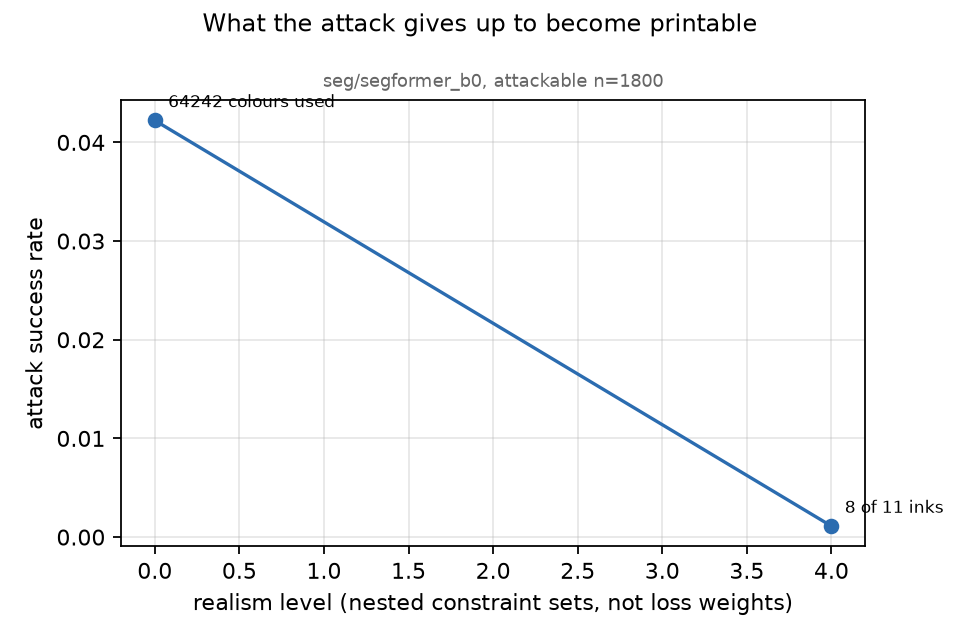}
  \caption{Attack success rate against realism constraint level at matched patch
  size, optimisation budget, supercell, victim and instance set. The subtitle
  records the victim and the attackable count, both held fixed.}
  \label{fig:curve}
\end{figure}

Holding patch size, optimisation budget, supercell, victim and instance set
fixed and varying only the constraint level, \asr{} on SegFormer-B0 falls from
$0.042$ at level~0 to $0.001$ at level~4 (\Cref{fig:curve}). Both runs use a
\SI{1.2}{\metre} patch, 3000 steps, supercell 8, batch 2 and the same 855
training instances; the only recorded settings that differ are the constraint
level, the anchor design it introduces, and the resulting degrees of freedom,
which fall from \num{196608} to $378$.

A second pair contrasts a \SI{2.0}{\metre} unconstrained patch against a
\SI{2.0}{\metre} level-4 patch, for which near-nadir \asr{} falls from $0.555$
to $0.223$. This pair is \emph{not} matched on optimisation budget: the
constrained run used 6000 steps against 3000 and a supercell of 16 against 8,
alongside the change in constraint level that reduces its degrees of freedom
from \num{196608} to \num{1386}. The reduction therefore cannot be attributed
to the loss of degrees of freedom alone, and no percentage is quoted for it.
It is reported because its direction is unambiguous in spite of the confound:
the constrained run received twice the optimisation budget and still reached
under half the attack rate.

A finer reading is not supported. An under-trained sweep at 400 optimisation
steps produces a non-monotonic curve peaking at level~2; the non-monotonicity
does not appear at 3000 steps. We therefore report the matched two-point
comparison and do not present a five-point curve at a budget that under-trains
the low-constraint levels, where the search space contains $196{,}608$ free
parameters.

\subsection{Attribution analysis}\label{sec:res-cam}
Coverage-based accounts of patch attacks assume the decal removes evidence from
the region it covers. An alternative account is that it introduces a salient
region drawing attribution away from the vehicle. The two make different
predictions under changes in patch size and placement.

\begin{figure}[htbp]
  \centering
  \includegraphics[
    width=\textwidth,
    height=0.75\textheight,
    keepaspectratio
  ]{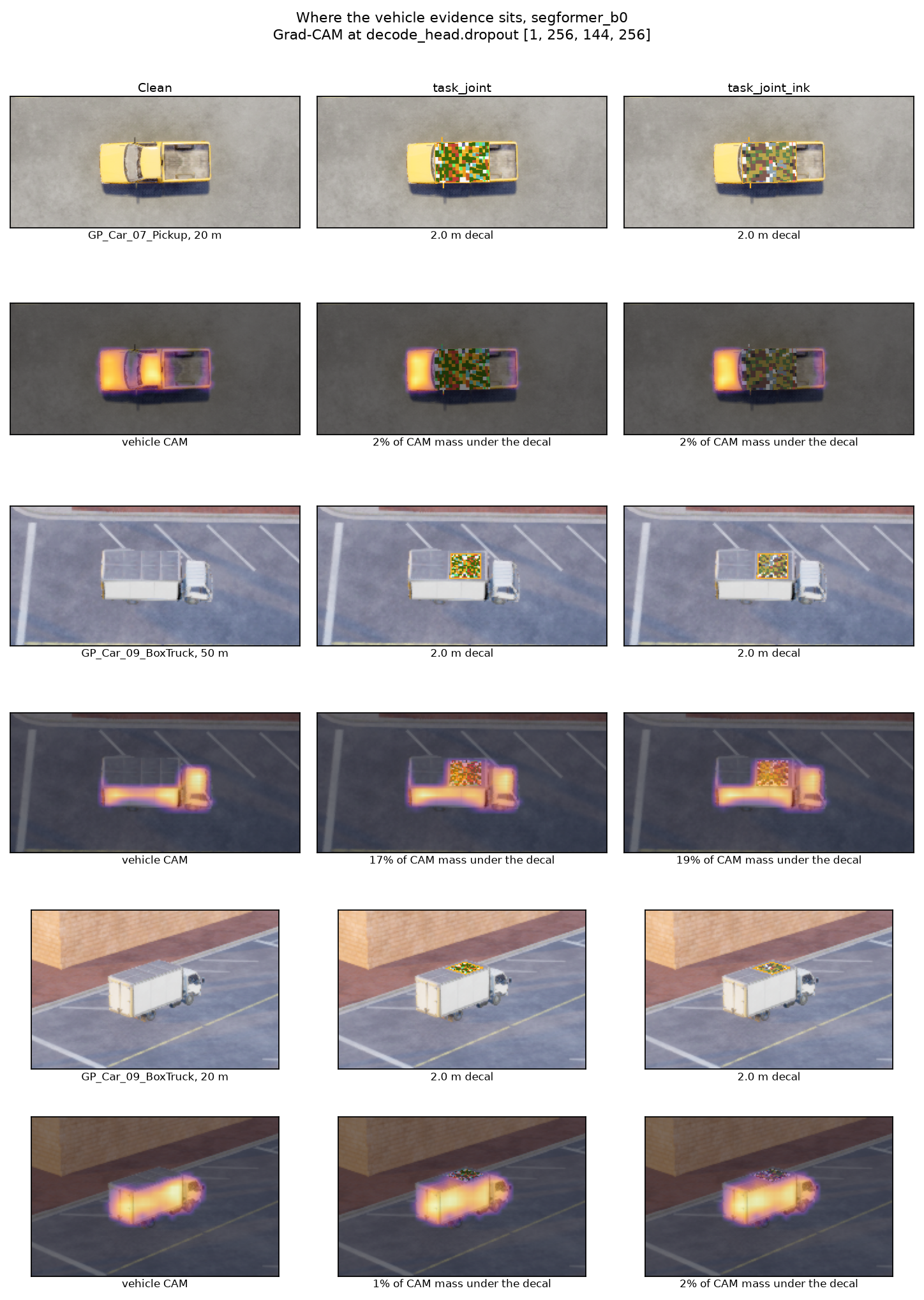}
  \caption{Grad-CAM attributions for the vehicle channel, SegFormer-B0, computed
  at the final 256-channel feature map. Annotations give the fraction of
  class-activation mass falling inside the decal footprint.}
  \label{fig:cam}
\end{figure}

\Cref{fig:cam} shows Grad-CAM attributions \cite{selvaraju2017grad} for the
vehicle channel. Under both ink sets the clean attribution covers the vehicle;
after the decal is applied, attribution does not relocate but is reduced within
the covered region, leaving between $0$ and $6$ per cent of class-activation
mass inside the decal footprint. On the box truck, where coverage is
approximately $0.18$, the attribution changes little. These observations are
consistent with the suppression account and do not support the diversion account
in this setting.

Attributions are reported for the segmentation victims only. Faster R-CNN
scores sampled proposals and YOLOv8n is accessed through a predictor interface,
so neither exposes a differentiable per-location vehicle response. For the two
dense detectors such a response is available, but the resulting maps were not
vehicle-localised: attribution concentrated away from the annotated instance,
and the fraction of mass inside the decal footprint was zero in every cell
tested, across several choices of target layer. Detection attributions are
therefore not reported and no mechanistic conclusion is drawn for the detection
arm. Whether this reflects the multi-level structure of the dense vehicle
response or an unsuitable target layer is unresolved.

\begin{figure}[htbp]
  \centering
  \includegraphics[
    width=\textwidth,
    height=0.75\textheight,
    keepaspectratio
  ]{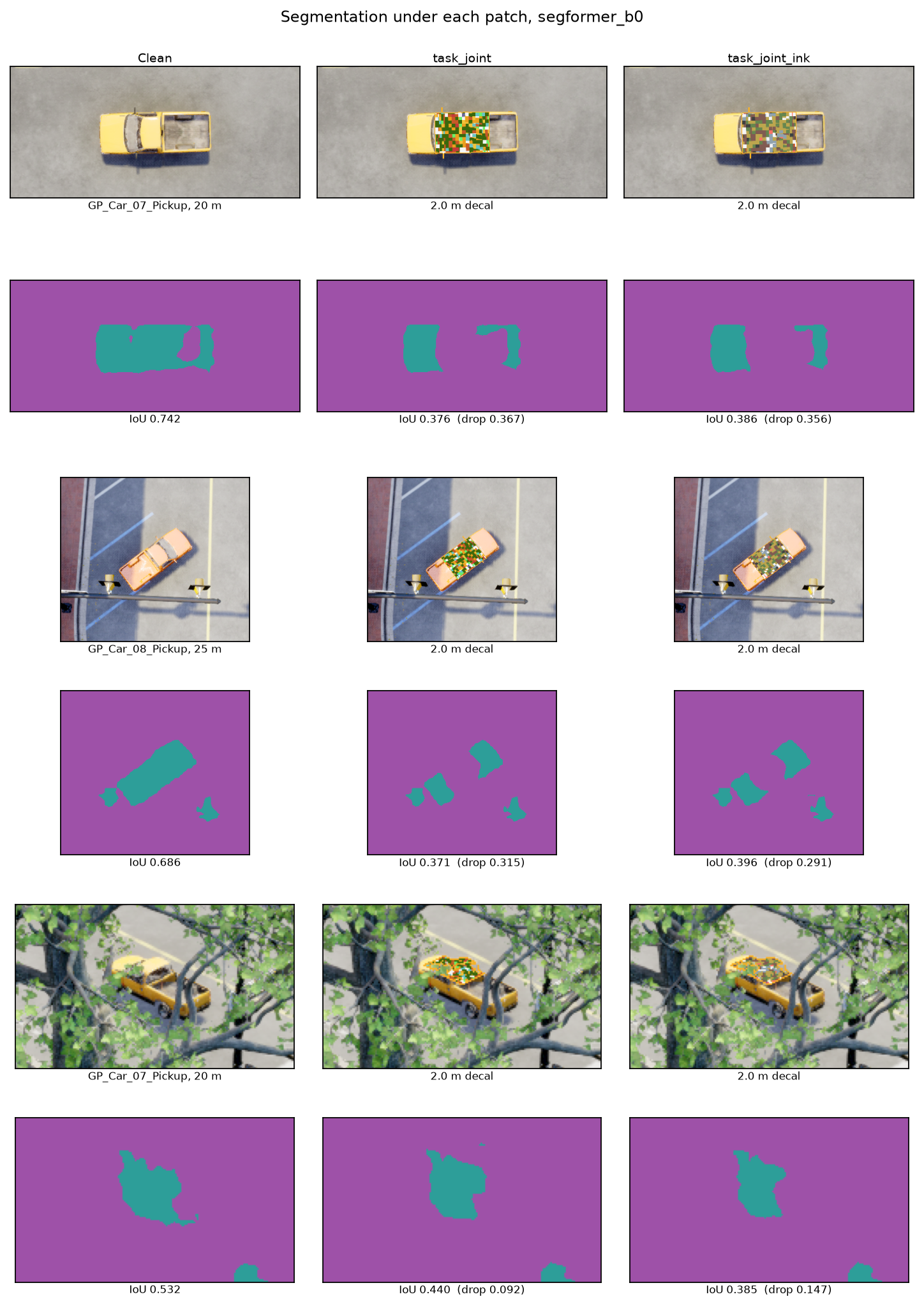}
  \caption{Semantic segmentation under each ink set, SegFormer-B0, on three
  instances where the attack succeeded. Columns are the clean frame, the
  colour-cube patch, and the measured-gamut patch; rows alternate input image
  and predicted vehicle map, annotated with intersection over union against
  ground truth and its drop from the clean prediction. Instances were drawn
  evenly across the nadir angle range from the 181 successes of the cube patch
  within a pool of 1800 attackable instances. This is a selected sample,
  included to show the failure mode rather than its frequency.}
  \label{fig:compare-seg-success}
\end{figure}

\Cref{fig:compare-seg} is drawn from the attackable pool without regard to
outcome, so most of its instances are ones where the attack did not succeed; it
shows the typical case. \Cref{fig:compare-seg-success} draws instead from the
instances the attack did flip, and shows what success looks like: the predicted
vehicle region shrinks, in two of the three instances fragmenting into
disconnected components, with intersection over union falling by between $0.09$
and $0.37$ across the panels shown. Selection is on the
\emph{cube} patch's successes, so the measured-gamut column is not selected in
its own favour, and in some such instances it recovers less of the vehicle than
the cube patch does. Both figures are selected samples and neither carries rate
information; the rates are in \Cref{tab:net}.

\begin{figure}[htbp]
  \centering
  \includegraphics[width=\linewidth]{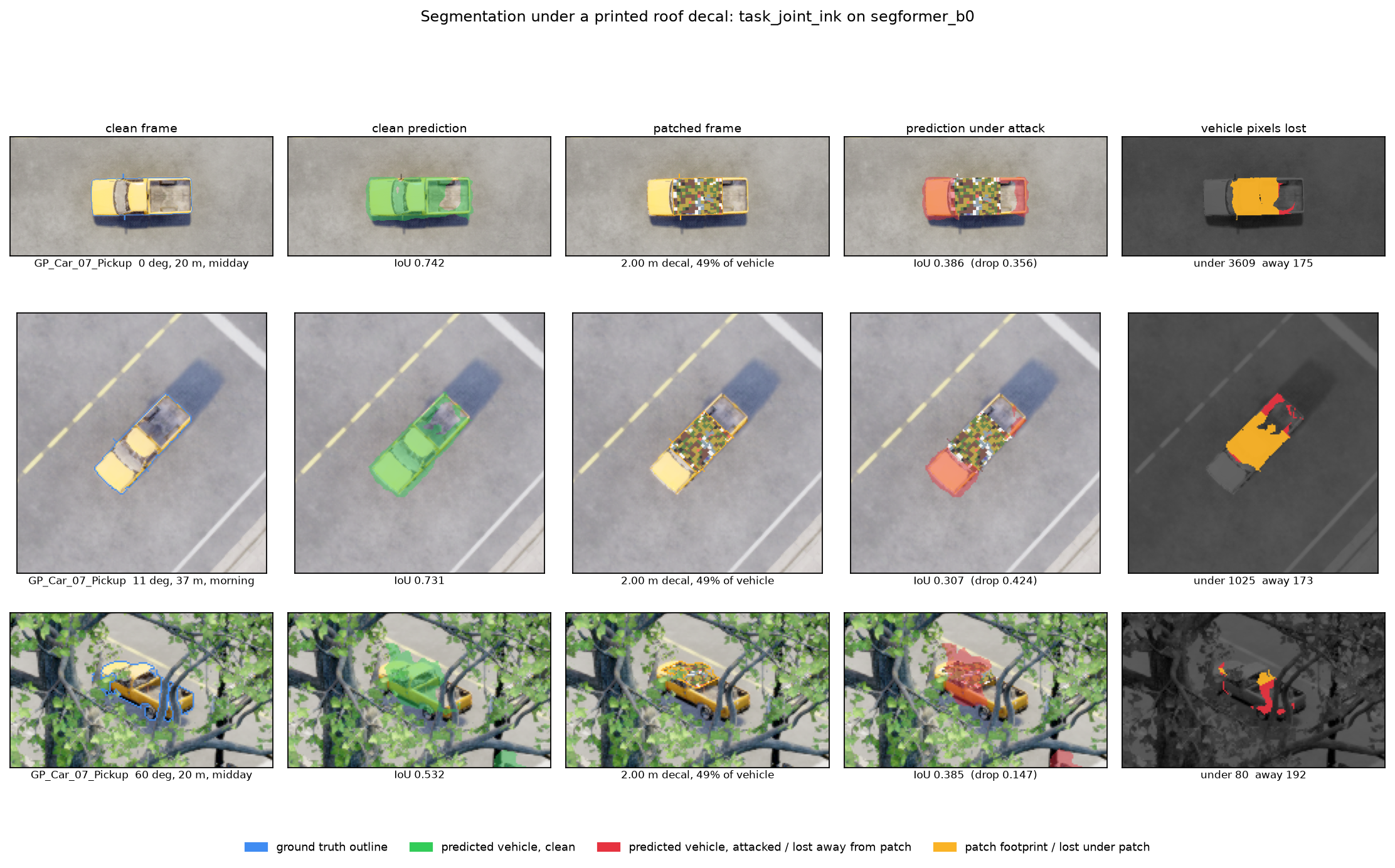}
  \caption{Per-instance segmentation detail for the measured-gamut patch on
  SegFormer-B0. Columns give (a) the clean frame with ground truth outlined,
  (b) the clean prediction, (c) the patched frame with the decal footprint
  outlined, (d) the prediction under attack, and (e) the flipped pixels
  separated into those under the decal and those away from it.}
  \label{fig:panel-seg}
\end{figure}

\Cref{fig:panel-seg} shows individual instances in more detail, separating the
pixels the attack flips into those beneath the decal and those away from it.

\subsection{Verification of the attack objective}\label{sec:res-objcheck}
In \Cref{fig:panel-seg,fig:compare-seg-success} most of the vehicle pixels the
attack removes lie beneath the decal. Over the full holdout on SegFormer-B0 the
measured-gamut patch removes $11.1$ per cent of visible vehicle pixels under
its footprint and $1.9$ per cent away from it. Suppression confined to the
footprint is what occlusion alone would produce, so it raises the question
whether the objective of \Cref{eq:segloss} is capable of more, or whether the
optimisation is failing. Three checks separate the two.

\paragraph{Formulation of the attack term.} \Cref{eq:segloss} hinges the
posterior, whose gradient with respect to the vehicle logit carries the factor
$p_{\text{veh}}(1-p_{\text{veh}})$. On SegFormer-B0 the median vehicle pixel
outside the footprint has log-odds $4.6$, a gradient factor near $10^{-2}$; on
UPerNet-ConvNeXt-T and UPerNet-Swin-T the medians are $6.9$ and $8.2$, and the
posterior clamp at $1-10^{-6}$ zeroes the gradient on up to $8.4$ per cent of
such pixels. \Cref{tab:objvariants} replaces the term with alternatives at
matched settings: the same hinge on the log-odds, which gives every unsuppressed
pixel a unit gradient; the log-odds hinge restricted to pixels outside the
footprint, so that occlusion earns nothing; and an untargeted term that pushes
every pixel in the frame off its true label, the formulation of the patch
literature. The level-4 patch was also reparametrised as one logit per anchor
region and ink, with a straight-through softmax, since the identity gradient of
the projection in \Cref{sec:method-realism} carries no information about the
snap to the nearest ink: fitted to four fixed frames without EOT, the original
parametrisation plateaued at a loss of $0.68$ within 100 steps with $19.5$ per
cent of vehicle pixels removed away from the footprint, against $0.43$ and
$28.2$ per cent under the reparametrisation. None of the alternatives improves
the held-out result. The log-odds forms match \Cref{eq:segloss} within
noise; restricting the term to pixels away from the footprint gives up the
occlusion without gaining anything elsewhere; and the untargeted term,
dominated by background far from the decal, suppresses nothing and raises
background predicted as vehicle from $0.2$ to $0.7$ per cent.

\begin{table}[htbp]
  \centering \small
  \caption{Alternative segmentation attack terms, SegFormer-B0, level~4 unless
  stated, \SI{2.0}{\metre} decal, 1000 optimisation steps, one seed each.
  Scored on a stride-5 subset of the holdout (507 instances, 368 attackable),
  so rates are comparable within this table but not with \Cref{tab:net}.
  ``Under'' and ``away'' are the shares of vehicle pixels removed beneath and
  outside the decal footprint.}
  \label{tab:objvariants}
  \begin{tabular}{lrrrr}
    \toprule
    Attack term & \asr{} & IoU drop & Under & Away \\
    \midrule
    Posterior hinge, \Cref{eq:segloss}         & 0.147 & $0.047$  & 11.3\% & 1.9\% \\
    Log-odds hinge                              & 0.136 & $0.037$  & 10.3\% & 1.9\% \\
    Log-odds hinge, per-region ink logits       & 0.139 & $0.040$  & 10.5\% & 2.0\% \\
    Log-odds hinge, away from footprint only    & 0.019 & $-0.031$ &  3.0\% & 1.9\% \\
    Untargeted, whole frame                     & 0.000 & $-0.041$ &  0.0\% & 0.2\% \\
    \midrule
    Posterior hinge, level~0, no print chain    & 0.171 & $0.080$  & 15.6\% & 2.2\% \\
    \bottomrule
  \end{tabular}
\end{table}

\begin{figure}[htbp]
  \centering
  \includegraphics[
    width=\textwidth,
    height=0.75\textheight,
    keepaspectratio
  ]{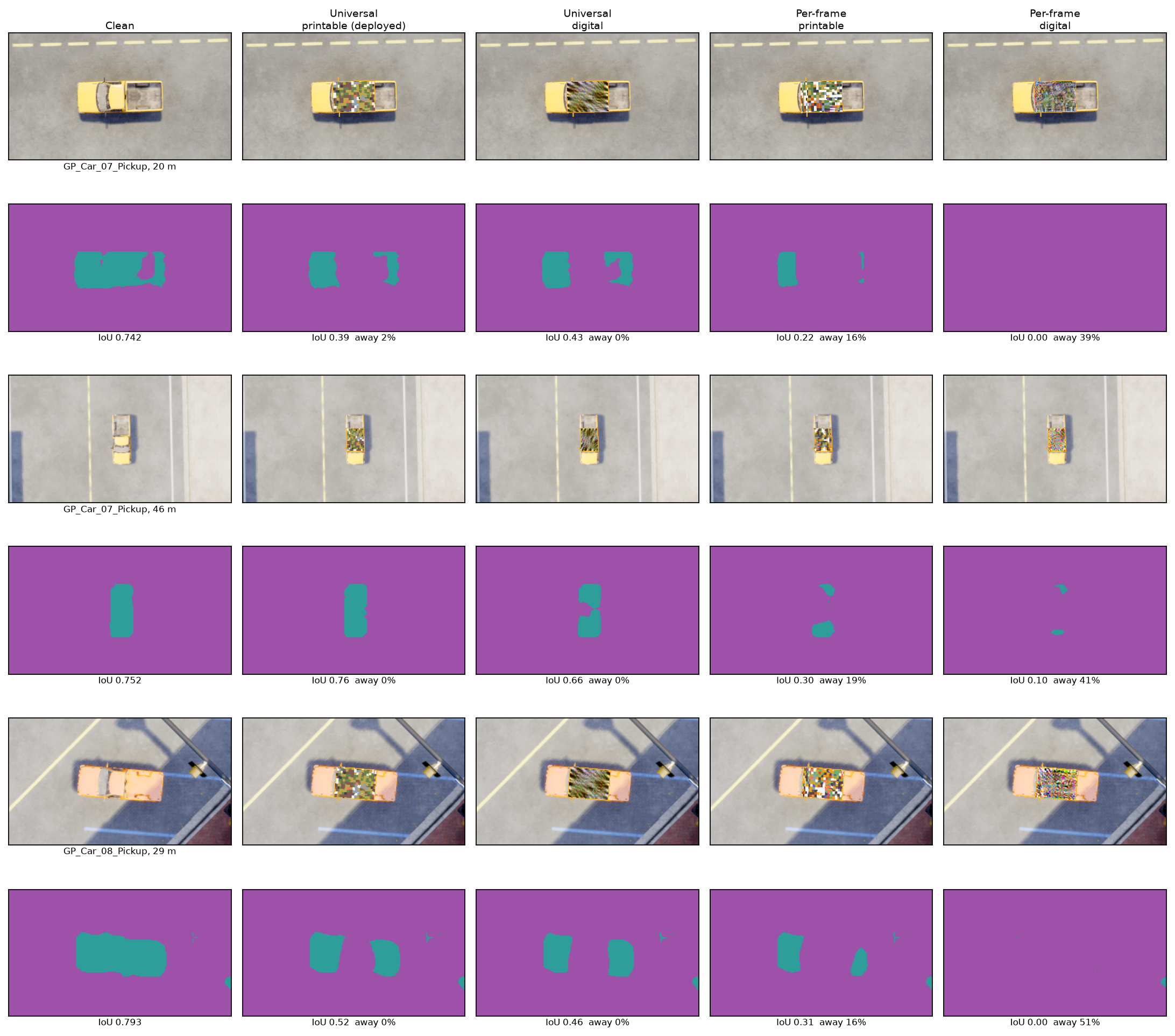}
  \caption{Segmentation of the same holdout instances under universal and
  per-frame patches, SegFormer-B0. Universal patches were optimised once over
  855 training instances; per-frame patches against the frame shown alone, with
  the same objective, footprint and placement and without EOT or the print
  chain. Rows alternate input, with the decal footprint outlined, and predicted
  vehicle map; ``away'' is the share of vehicle pixels removed outside the
  footprint. The instances are evenly spaced over the 24 of
  \Cref{tab:objcheck}, which were chosen without reference to any attack
  outcome.}
  \label{fig:objcheck}
\end{figure}

\paragraph{Per-frame against universal patches.} The decisive test holds the
objective fixed and varies only what the patch must serve. For 24 attackable
nadir holdout instances, chosen at even spacing without reference to any attack
outcome, a patch was optimised against each frame alone, with
\Cref{eq:segloss}, the same \SI{2.0}{\metre} footprint and the same roof
placement, for 300 steps without EOT or the print chain, once under the level-4
constraint set and once as free pixels. \Cref{tab:objcheck} compares these with
the deployed measured-gamut patch and with a universal free-pixel patch
optimised over the 855 training instances for 3000 steps without the print
chain. Fitted to one frame, the objective removes the vehicle well beyond the
decal: $47.0$ per cent of vehicle pixels away from the footprint for the
free-pixel patch, never less than $38.6$ per cent on any instance, and $26.2$
per cent under the printability constraints. Required to serve every vehicle at
once, the same objective removes $3.0$ and $6.5$ per cent. \Cref{fig:objcheck}
shows three of the instances in the layout conventional in the patch
literature.

\begin{table}[htbp]
  \centering \small
  \caption{The objective of \Cref{eq:segloss} on one frame at a time and across
  the fleet, SegFormer-B0, 24 attackable nadir holdout instances, all
  \SI{2.0}{\metre} footprints. Per-frame patches are optimised against the
  instance they are scored on and are an upper bound, not an attack.}
  \label{tab:objcheck}
  \begin{tabular}{lrrrr}
    \toprule
    Patch & \asr{} & IoU drop & Under & Away \\
    \midrule
    Universal, printable (deployed) & 0.417 & 0.195 & 25.2\% &  3.0\% \\
    Universal, free pixels          & 0.583 & 0.296 & 34.1\% &  6.5\% \\
    Per-frame, printable            & 1.000 & 0.567 & 48.0\% & 26.2\% \\
    Per-frame, free pixels          & 1.000 & 0.755 & 48.4\% & 47.0\% \\
    \bottomrule
  \end{tabular}
\end{table}

\paragraph{Universality, not generalisation.} The limit is not a gap between
training and held-out vehicles. The universal free-pixel patch removes $3.6$
per cent of vehicle pixels away from the footprint over the full holdout, at
\asr{} $0.265$ against $0.127$ for the deployed patch and $0.082$ for a grey
square of the same size, which removes $0.5$ per cent away from its footprint.
A 1000-step free-pixel patch scored on its own training instances removes $3.3$
per cent away from the footprint, against $2.2$ per cent held out.

The objective is therefore capable of suppressing the vehicle far beyond the
decal, and the optimisation reaches that regime whenever the patch is fitted to
one frame. What confines the effect of the reported patches to their footprint
is the requirement that one decal serve every vehicle, altitude and placement.
Qualitative figures of patch attacks on aerial segmentation that show
disruption far beyond the patch should accordingly be compared only at matched
universality.

\begin{figure}[htbp]
  \centering
  \includegraphics[width=\linewidth]{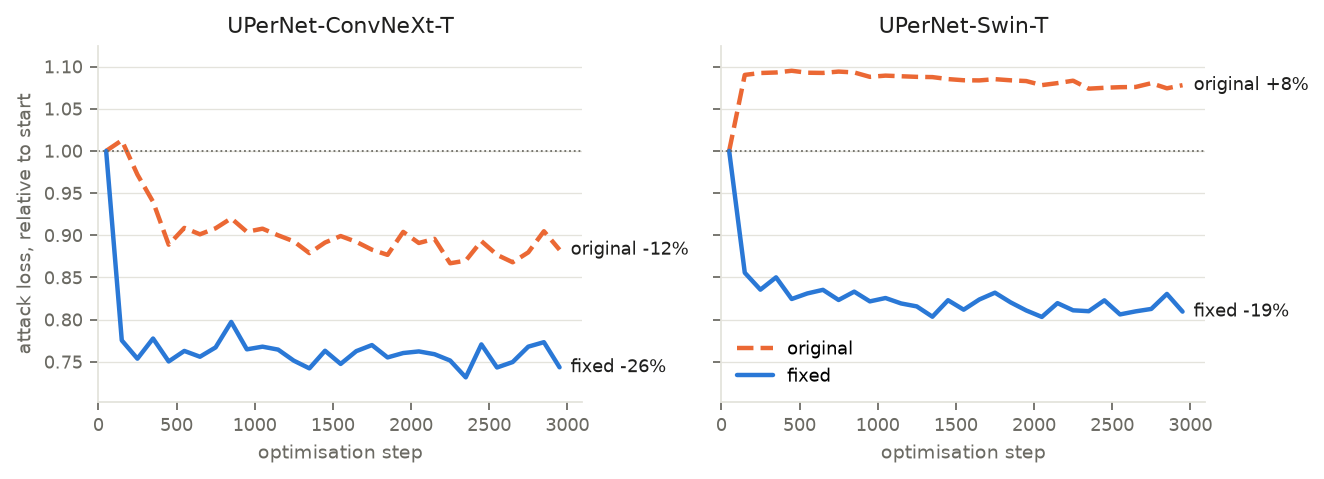}
  \caption{Attack loss during white-box optimisation against the UPerNet pair,
  in 100-step means. The original and fixed runs optimise attack terms on
  different scales, so each curve is indexed to its own first block; values
  below 1 are descending. The original Swin-T run diverges.}
  \label{fig:wbloss}
\end{figure}

\begin{figure}[htbp]
  \centering
  \includegraphics[width=\linewidth]{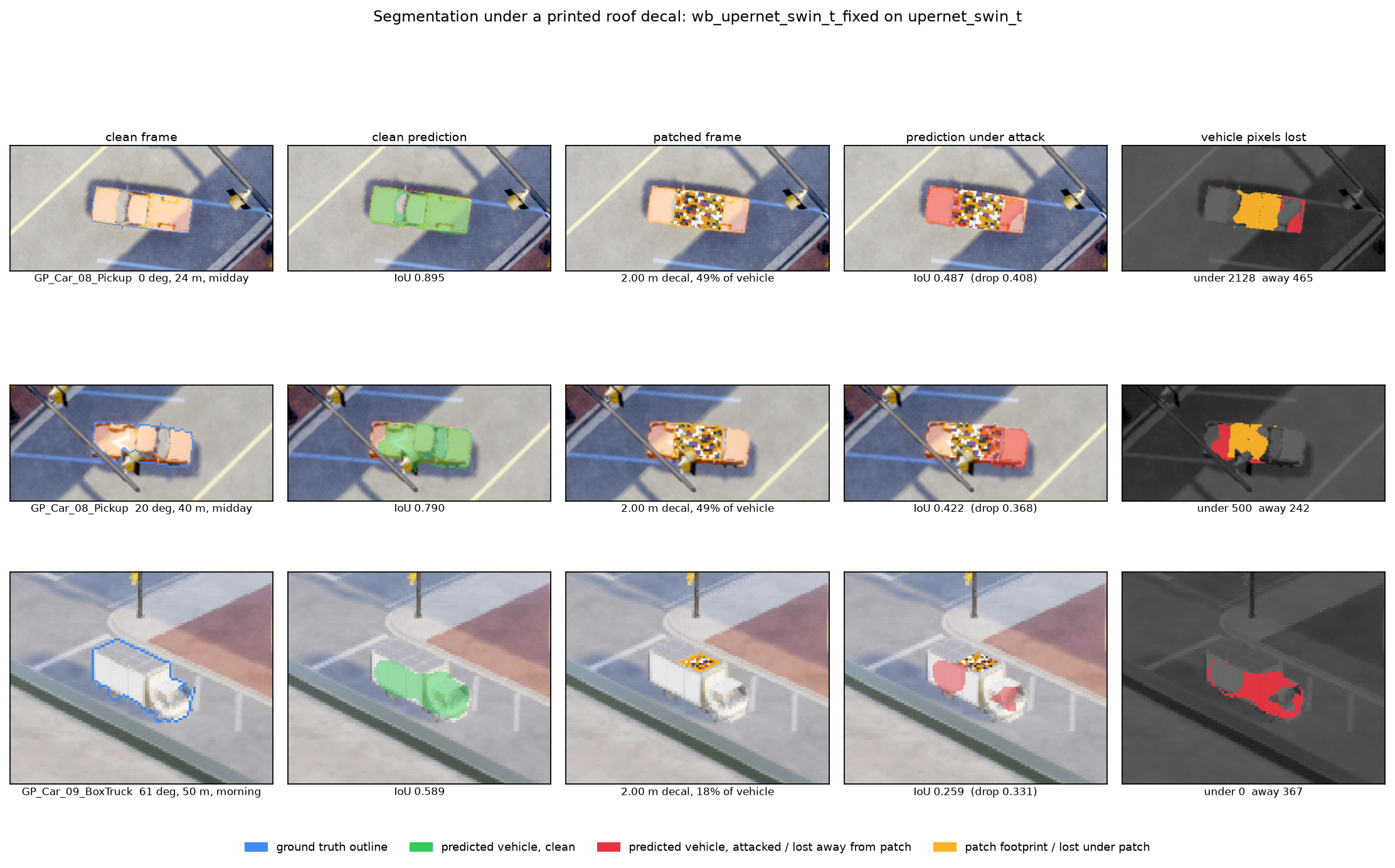}
  \caption{The repaired white-box attack on UPerNet-Swin-T. Columns as in
  \Cref{fig:panel-seg}. The three instances are successes, drawn evenly across
  the viewing-angle range from 1879 attackable instances; this is a selected
  sample, and the rate is in \Cref{tab:wb}.}
  \label{fig:wbpanel}
\end{figure}

\paragraph{A diverging white-box run.} The checks also resolved a failure the
optimisation had hidden. UPerNet-ConvNeXt-T and UPerNet-Swin-T share a decoder
and differ in whether the backbone attends, so attacking each white box under
identical settings isolates the effect of local windowed attention. Under the
original attack term and parametrisation, the Swin run did not converge: its
loss rose by 8 per cent over 3000 steps while the ConvNeXt run's fell by 12,
and its patch reached \asr{} $0.005$, no better than the unoptimised control.
Read at face value this would suggest that windowed attention confers
immunity. Both runs were repeated with the log-odds term and the per-region
ink parametrisation, all other settings unchanged
(\SI{2.0}{\metre}, level~4, supercell 16, 3000 steps, print chain on). Both
losses then descend (\Cref{fig:wbloss}), and the Swin patch reaches a net of
$+0.056$ (\Cref{tab:wb}). Swin-T is therefore attackable white box, and its
earlier score measured the optimiser rather than the architecture. It remains
less attackable than ConvNeXt-T, at under two fifths of its net under
identical treatment; this is a single run per model, so the ratio is
indicative rather than resolved. The fixes also raised the ConvNeXt-T net from
$+0.115$ to $+0.152$. Each patch is stronger on its own victim than on the
other member of the pair: the ConvNeXt-T patch reaches $+0.032$ on Swin-T by
transfer and the Swin-T patch $+0.060$ on ConvNeXt-T. \Cref{fig:wbpanel} shows
instances the repaired Swin-T attack flips.

\begin{table}[htbp]
  \centering \small
  \caption{White-box attacks on the UPerNet pair, full 2534-instance holdout,
  one run each. ``Original'' uses \Cref{eq:segloss} and the projected
  parametrisation; ``fixed'' uses the log-odds hinge and per-region ink
  logits. The control is the unoptimised reference decal at the same size.}
  \label{tab:wb}
  \begin{tabular}{llrrrrr}
    \toprule
    Victim & Run & Attackable & \asr{} & Control & $\net$ & Nadir \asr{} \\
    \midrule
    UPerNet-ConvNeXt-T & original & 2080 & 0.125 & 0.010 & $+0.115$ & 0.164 \\
                       & fixed    & 2080 & 0.162 & 0.010 & $+0.152$ & 0.184 \\
    \midrule
    UPerNet-Swin-T     & original & 1879 & 0.005 & 0.003 & $+0.002$ & 0.000 \\
                       & fixed    & 1879 & 0.059 & 0.003 & $+0.056$ & 0.092 \\
    \bottomrule
  \end{tabular}
\end{table}

\subsection{Detection arm and the magnitude of the control}\label{sec:res-det}
No net attack success in the detection arm exceeds $+0.026$ for either ink set,
against segmentation nets reaching $+0.596$ (\Cref{tab:net}). Raw detection
\asr{} ranges from $0.029$ to $0.068$, while the corresponding size-matched
controls range from $0.011$ to $0.049$: between one half and nine tenths of the
raw figure is attributable to the presence of an applied object rather than to
its optimised content. FCOS under the measured ink set is the clearest case, at
$0.053$ raw against a $0.049$ control, giving $\net = +0.004$.

This observation extends beyond the present study. A detection attack success
rate reported without a size-matched control may be dominated by the control and
is difficult to interpret in isolation.

\begin{figure}[htbp]
  \centering
  \includegraphics[
    width=\textwidth,
    height=0.75\textheight,
    keepaspectratio
  ]{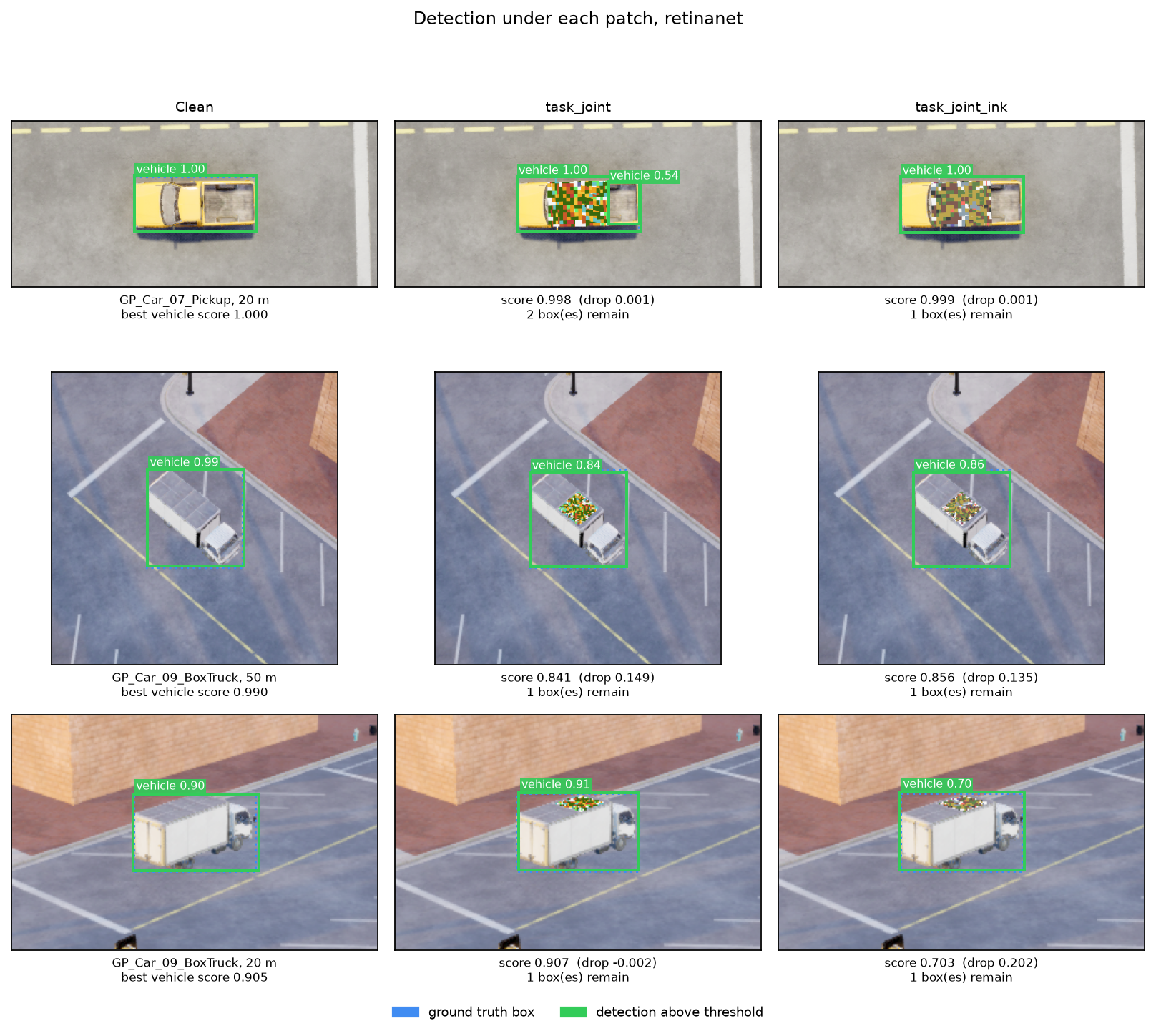}
  \caption{Object detection under each ink set, RetinaNet. The ground-truth box
  is shown dotted and detections above threshold solid. Confidence and its
  change relative to the clean frame are given beneath each panel.}
  \label{fig:compare-det}
\end{figure}

\begin{figure}[htbp]
  \centering
  \includegraphics[
    width=\textwidth,
    height=0.75\textheight,
    keepaspectratio
  ]{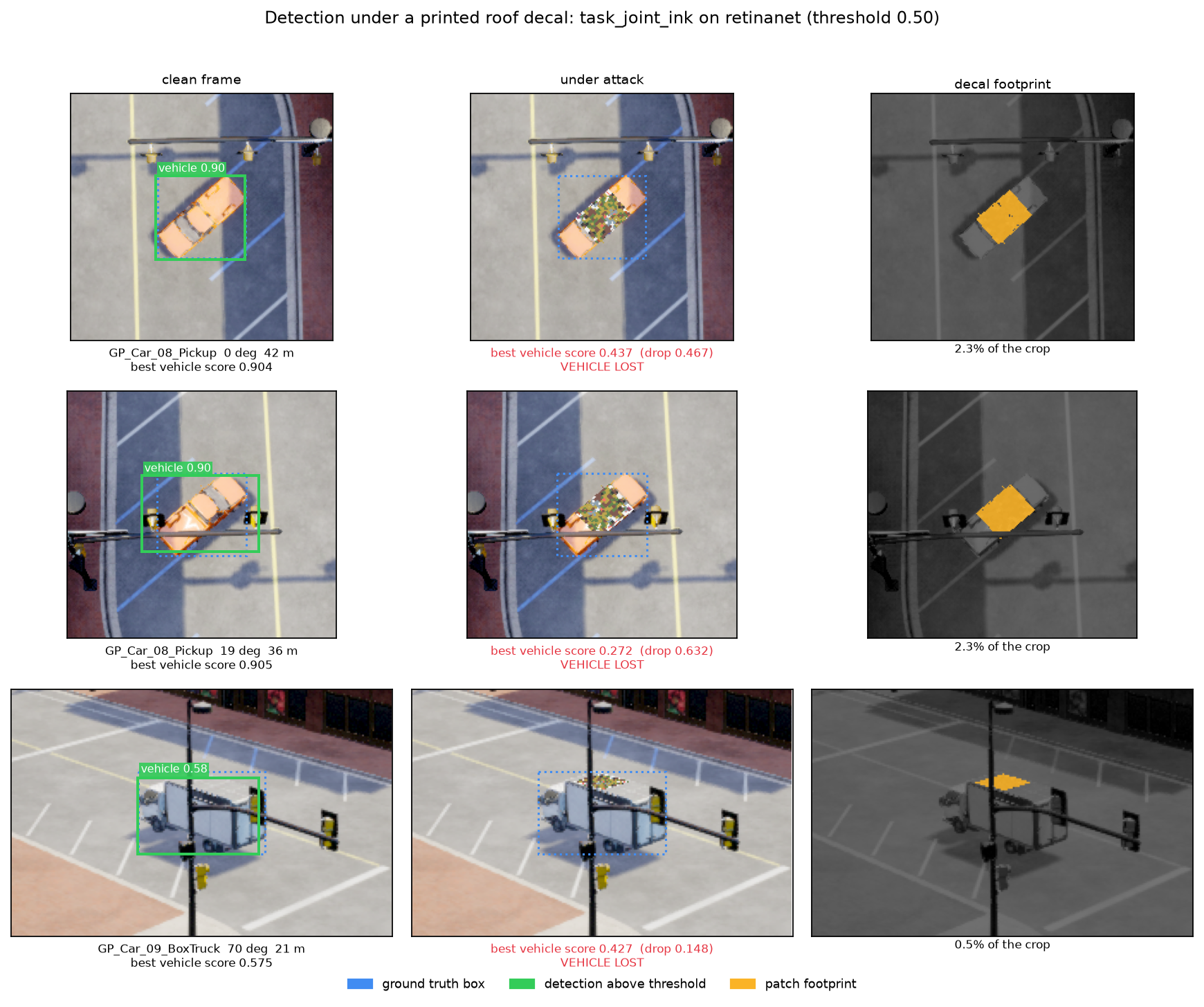}
  \caption{Per-instance detection detail for the measured-gamut patch on
  RetinaNet. Columns give (a) the clean frame, (b) the frame under attack, and
  (c) the decal footprint. Ground-truth boxes are dotted and detections above
  threshold solid; the note beneath each attacked frame states whether any box
  survived.}
  \label{fig:panel-det}
\end{figure}

\Cref{fig:compare-det} shows the qualitative effect of each ink set on matched
instances, and \Cref{fig:panel-det} separates the clean frame, the attacked
frame and the decal footprint for individual instances. In most instances the
vehicle box survives under both patches with a small reduction in confidence,
which is the per-instance form of the aggregate figures above.

\begin{figure}[htbp]
  \centering
  \includegraphics[
    width=\textwidth,
    height=0.75\textheight,
    keepaspectratio
  ]{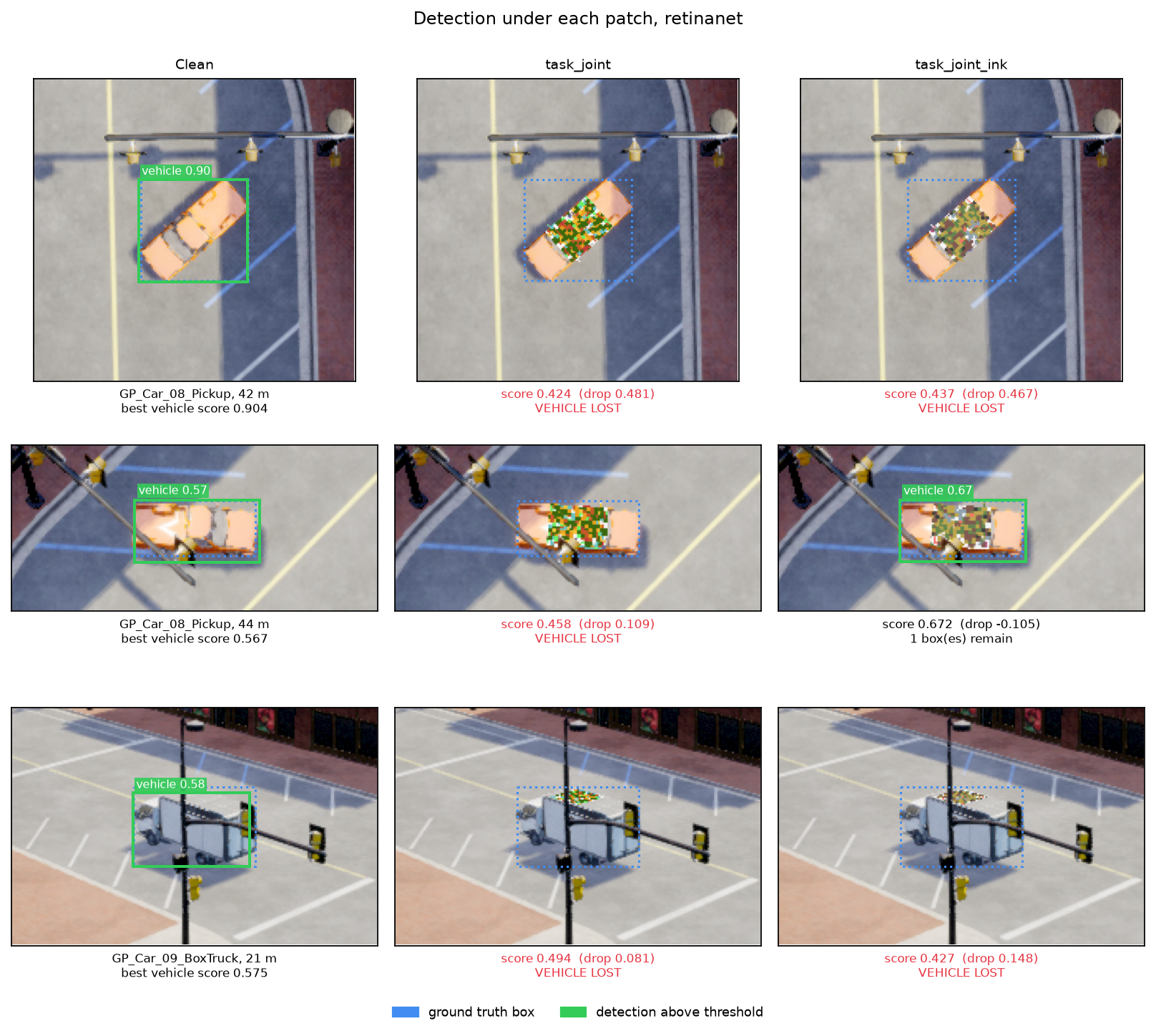}
  \caption{Object detection under each ink set, RetinaNet, on three instances
  where the attack succeeded. Ground-truth boxes are dotted and detections
  above threshold solid; the best vehicle score and its drop from the clean
  frame are given beneath each panel. Instances were drawn evenly across the
  nadir angle range from the 121 successes of the cube patch within a pool of
  2286 attackable instances. This is a selected sample, included to show the
  failure mode rather than its frequency.}
  \label{fig:compare-det-success}
\end{figure}

\Cref{fig:compare-det-success} shows the minority of instances in which the box
is lost. Success here is a confidence reduction that crosses the detection
threshold rather than a collapse of the response: the scores that survive
suppression lie between $0.42$ and $0.49$ against clean scores of $0.57$ to
$0.90$. The second row is informative in the other direction, being an instance
where the cube patch removed the box and the measured-gamut patch left it
standing at a \emph{higher} score than the clean frame. That is the
per-instance form of the detection result in \Cref{sec:res-seeds}, where the
cube patch is the stronger of the two on this arm.

\subsection{Transfer across viewpoint, altitude and illumination}\label{sec:res-envelope}
\begin{figure}[htbp]
  \centering
  \begin{subfigure}[b]{0.49\linewidth}
    \includegraphics[width=\linewidth]{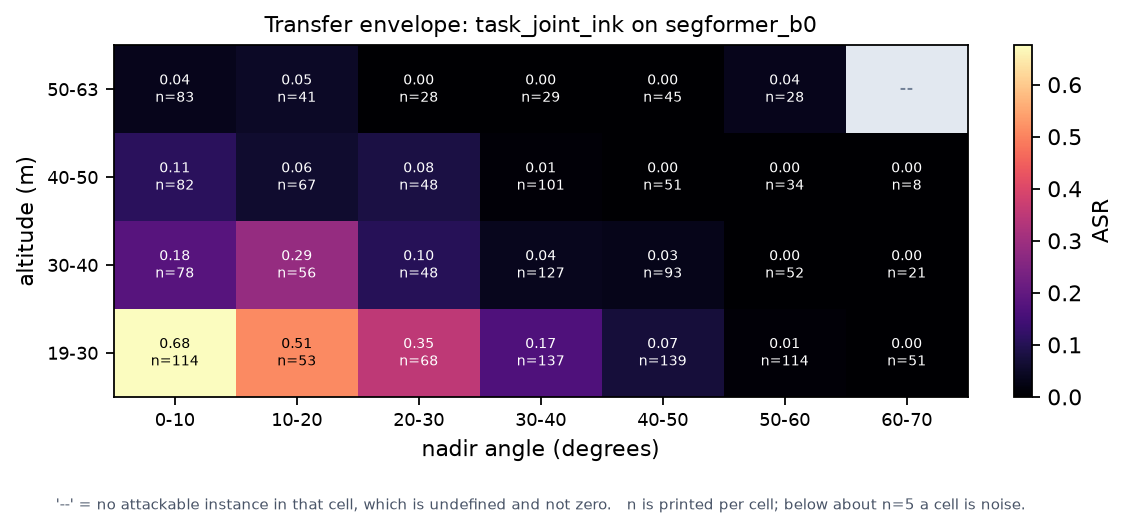}
    \caption{Semantic segmentation, SegFormer-B0}
    \label{fig:surface-seg}
  \end{subfigure}\hfill
  \begin{subfigure}[b]{0.49\linewidth}
    \includegraphics[width=\linewidth]{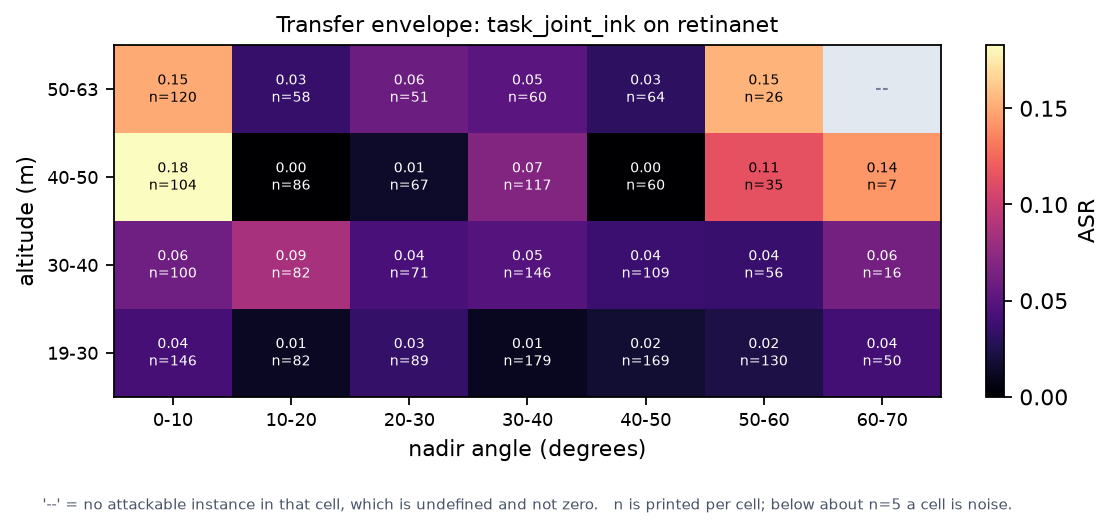}
    \caption{Object detection, RetinaNet}
    \label{fig:surface-det}
  \end{subfigure}
  \caption{Attack success rate over nadir angle and altitude for the
  measured-gamut patch. Each cell gives \asr{} and the number of attackable
  instances; cells without an attackable instance are marked undefined.}
  \label{fig:surface}
\end{figure}

\Cref{fig:surface} presents attack success jointly over nadir angle and altitude
for the measured-gamut patch. Cells are annotated with the number of attackable
instances; cells containing none are marked undefined rather than zero, per
\Cref{eq:asr}.

\begin{figure}[htbp]
  \centering
  \includegraphics[width=0.70\linewidth]{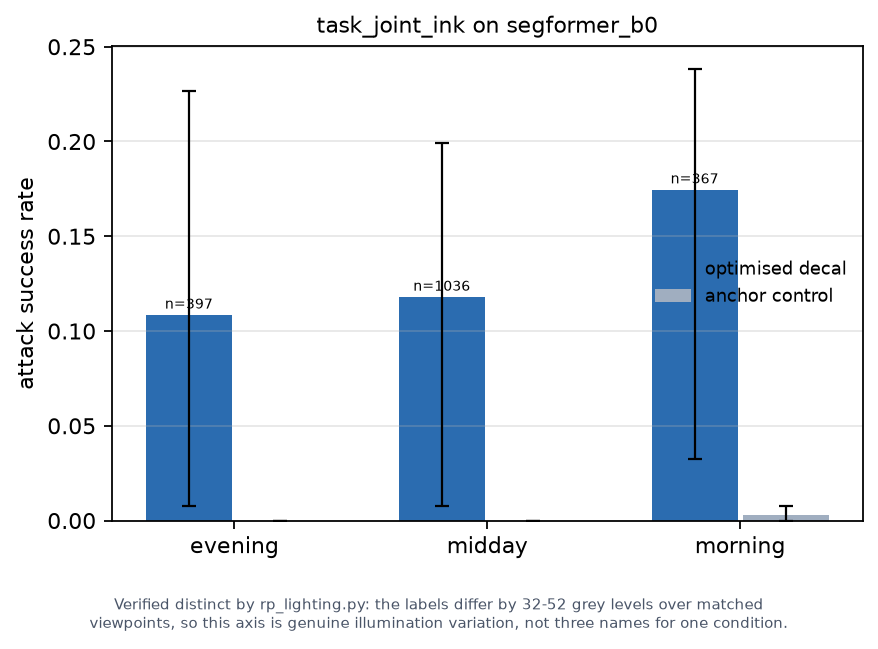}
  \caption{Attack success rate by illumination condition for the measured-gamut
  patch on SegFormer-B0, with vehicle-level confidence intervals.}
  \label{fig:lighting}
\end{figure}

Attack success decreases monotonically with nadir angle, from $0.289$ within $10$
degrees of nadir to $0.000$ beyond $60$ degrees, and monotonically with
altitude, from $0.691$ at $19$--$30$ metres to $0.036$ at $50$--$63$ metres
(\Cref{tab:envelope}). Variation across illumination conditions is smaller, from
$0.108$ to $0.174$ (\Cref{fig:lighting}), which suggests that viewing geometry
rather than illumination is the dominant factor over the range sampled. Both
trends are consistent with \Cref{eq:texels}: increasing altitude raises $g$ and
therefore $\rho$, moving more of the patch's structure above the channel cutoff.

\begin{figure}[htbp]
  \centering
  \begin{subfigure}[b]{0.49\linewidth}
    \includegraphics[width=\linewidth]{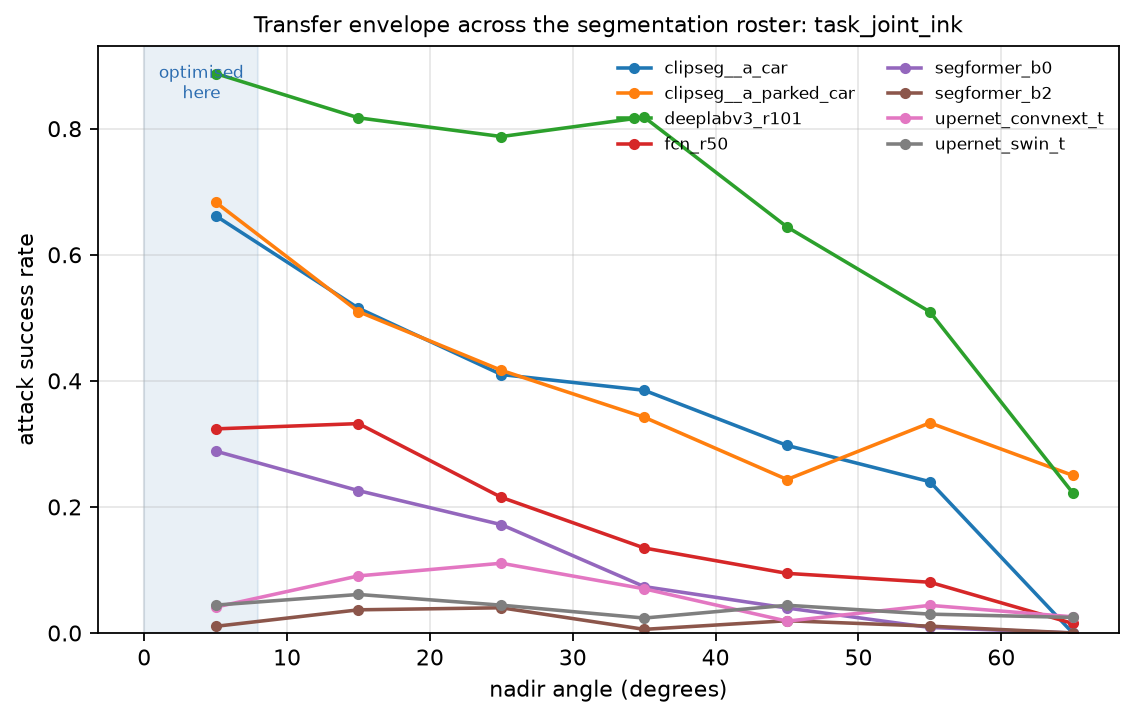}
    \caption{Segmentation roster}
    \label{fig:roster-seg}
  \end{subfigure}\hfill
  \begin{subfigure}[b]{0.49\linewidth}
    \includegraphics[width=\linewidth]{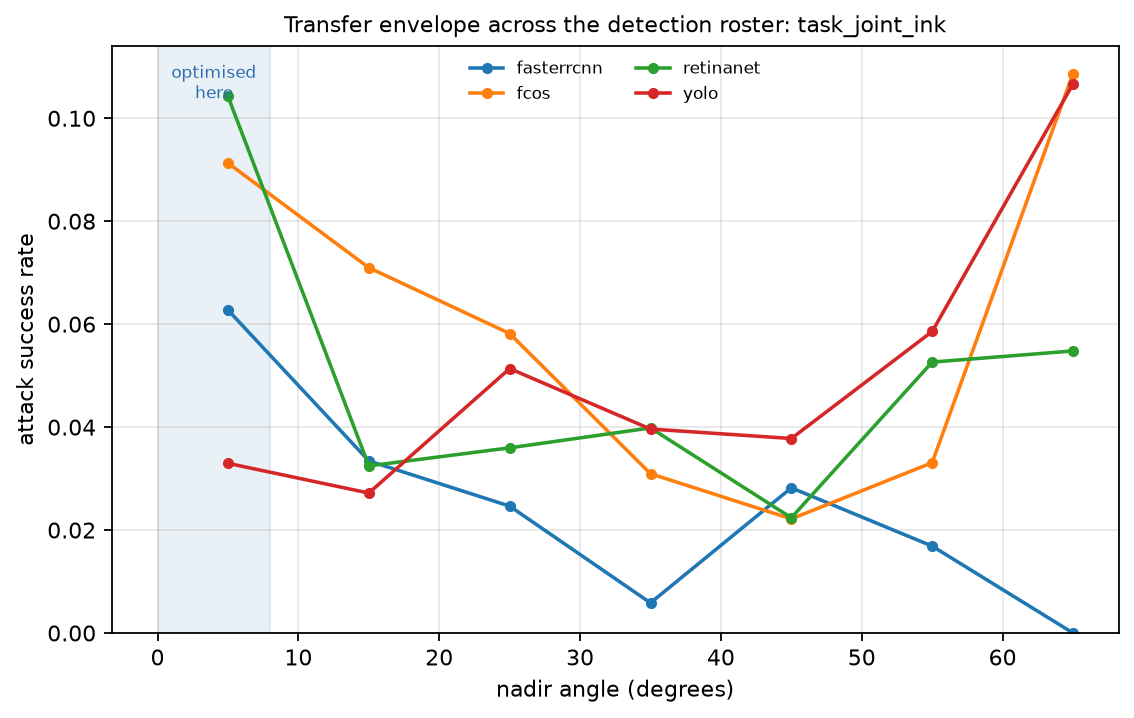}
    \caption{Detection roster}
    \label{fig:roster-det}
  \end{subfigure}
  \caption{Attack success against nadir angle for the measured-gamut patch
  across the full victim roster. The shaded band marks the near-nadir range the
  patch was optimised at.}
  \label{fig:roster}
\end{figure}

\Cref{fig:roster} places every victim on one axis. On the segmentation roster
the decline with nadir angle is monotonic for the victims with appreciable
attack success---DeepLabv3-R101, FCN-R50, SegFormer-B0 and both CLIPSeg
prompts---and DeepLabv3-R101 is the uppermost curve at every angle below
\SI{60}{\degree}, consistent with its anomalous attackable fraction noted in
\Cref{sec:limits}. For UPerNet-ConvNeXt-T and SegFormer-B2 the rate stays below
$0.12$ at every angle and varies non-monotonically; this is consistent with the
near-zero nets those victims record in \Cref{tab:net} and is not read here as a
trend. The spread between victims at a fixed angle, $0.01$ to $0.88$ within the
optimised band, exceeds the decline of any single victim across any
\SI{20}{\degree} interval, which is why a single-victim envelope is reported
separately in \Cref{tab:envelope} rather than averaged across the roster.

The detection roster shows no comparable structure. All four detectors remain
between $0.00$ and $0.11$ at every angle, a range that overlaps the
size-matched controls of \Cref{sec:res-det}, and FCOS and YOLOv8n rise rather
than fall beyond \SI{55}{\degree}. These curves are therefore reported as
showing no resolvable dependence on viewing angle, rather than as a shallower
form of the segmentation trend.

\begin{table}[htbp]
  \centering \small
  \caption{Transfer envelope for the measured-gamut patch on SegFormer-B0.
  Altitude bins are restricted to near-nadir frames so that viewing angle and
  altitude are not confounded; ``att.'' is the number of attackable instances.}
  \label{tab:envelope}
  \begin{tabular}{lrr @{\hspace{1.6em}} lrr @{\hspace{1.6em}} lrr}
    \toprule
    Nadir angle & att. & \asr{} & Altitude & att. & \asr{} & Illumination & att. & \asr{} \\
    \midrule
    \SIrange{0}{10}{\degree}  & 357 & 0.289 & \SIrange{19}{30}{\metre} & 68 & 0.691 & morning & 367 & 0.174 \\
    \SIrange{10}{20}{\degree} & 217 & 0.226 & \SIrange{30}{40}{\metre} & 55 & 0.182 & midday  & 1036 & 0.118 \\
    \SIrange{20}{30}{\degree} & 192 & 0.172 & \SIrange{40}{50}{\metre} & 60 & 0.067 & evening & 397 & 0.108 \\
    \SIrange{30}{40}{\degree} & 394 & 0.074 & \SIrange{50}{63}{\metre} & 55 & 0.036 & & & \\
    \SIrange{40}{50}{\degree} & 328 & 0.040 & & & & & & \\
    \SIrange{50}{60}{\degree} & 228 & 0.009 & & & & & & \\
    \SIrange{60}{70}{\degree} &  80 & 0.000 & & & & & & \\
    \bottomrule
  \end{tabular}
\end{table}

\begin{figure}[htbp]
  \centering
  \begin{subfigure}[b]{0.49\linewidth}
    \includegraphics[width=\linewidth]{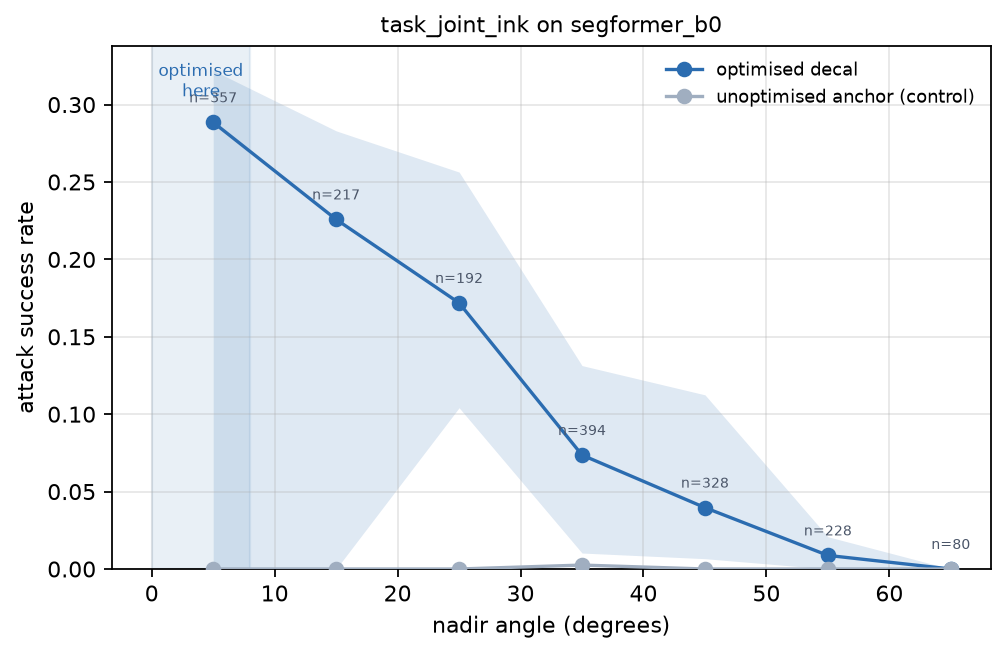}
    \caption{Nadir angle}
    \label{fig:marg-angle}
  \end{subfigure}\hfill
  \begin{subfigure}[b]{0.49\linewidth}
    \includegraphics[width=\linewidth]{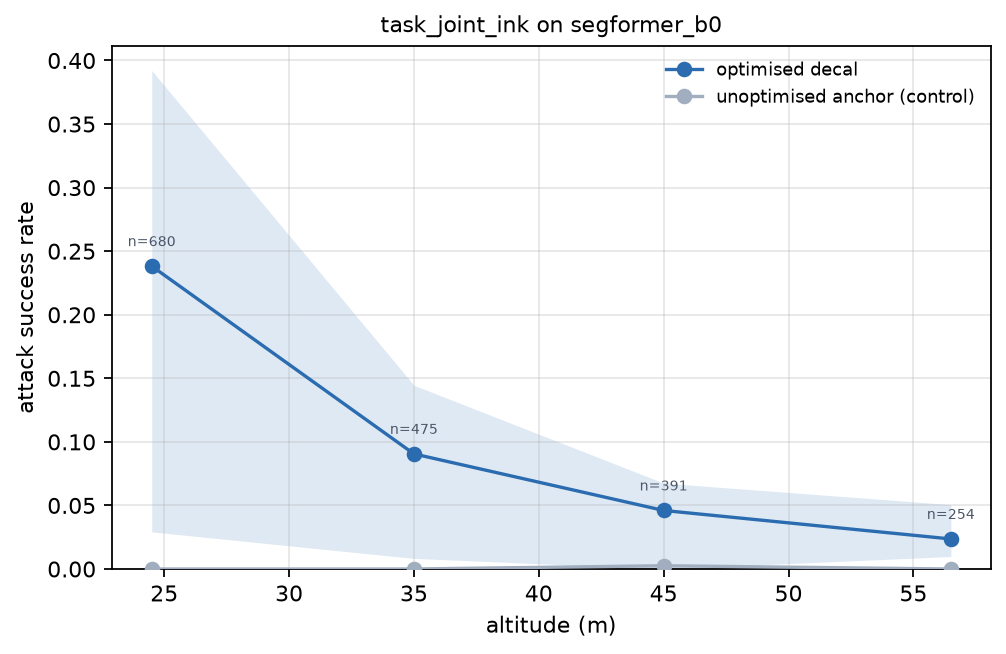}
    \caption{Altitude}
    \label{fig:marg-alt}
  \end{subfigure}
  \caption{Attack success against each geometric axis for the measured-gamut
  patch on SegFormer-B0, with the unoptimised size-matched control and
  vehicle-level confidence bands. Attackable counts are printed per bin.}
  \label{fig:marginals}
\end{figure}

\Cref{fig:marginals} gives the two geometric axes separately, with the
size-matched control overlaid and vehicle-level confidence bands. Three points
follow. The control is flat and indistinguishable from zero across both axes,
so the decline belongs to the optimised decal rather than to the presence of an
applied object. The bands are widest where the three held-out vehicles disagree
most, which is not where the attackable counts are smallest: the widest interval
on the altitude axis falls in the lowest bin, which has the largest count at
$n = \num{680}$. With three vehicles, one cluster-bootstrap resample in nine is
degenerate, so these intervals are conservative and are best read as bounds on
what a three-vehicle holdout can establish rather than as sampling error on the
point estimate. Finally, the altitude panel here marginalises over all nadir
angles and therefore lies below the altitude column of \Cref{tab:envelope},
which restricts to near-nadir frames so that the two geometric factors are not
confounded; the two are consistent, not conflicting.

\begin{figure}[htbp]
  \centering
  \begin{subfigure}[b]{0.49\linewidth}
    \includegraphics[width=\linewidth]{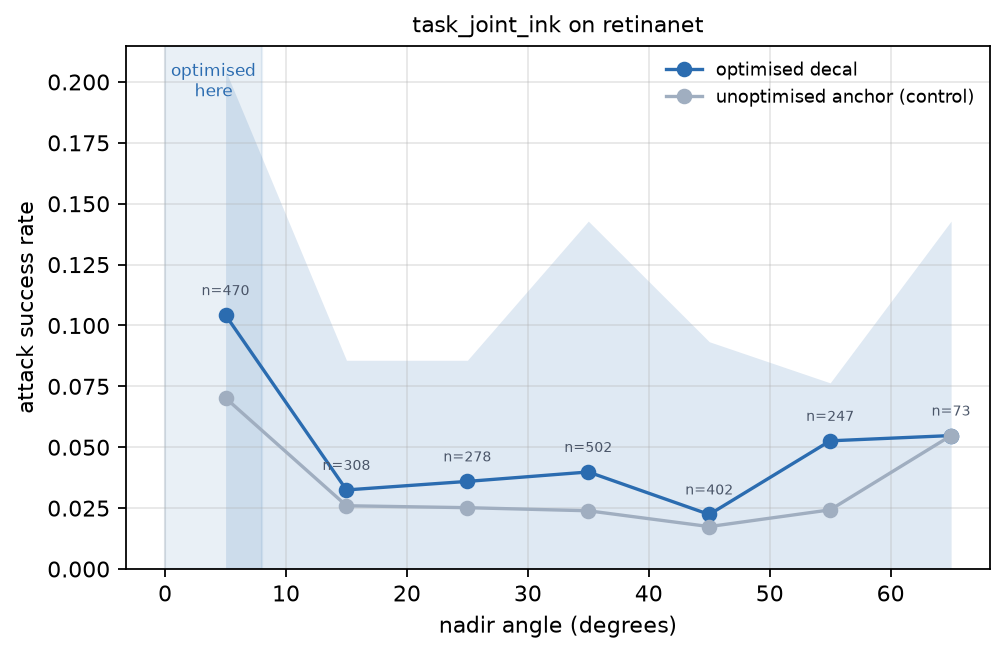}
    \caption{Nadir angle}
    \label{fig:margdet-angle}
  \end{subfigure}\hfill
  \begin{subfigure}[b]{0.49\linewidth}
    \includegraphics[width=\linewidth]{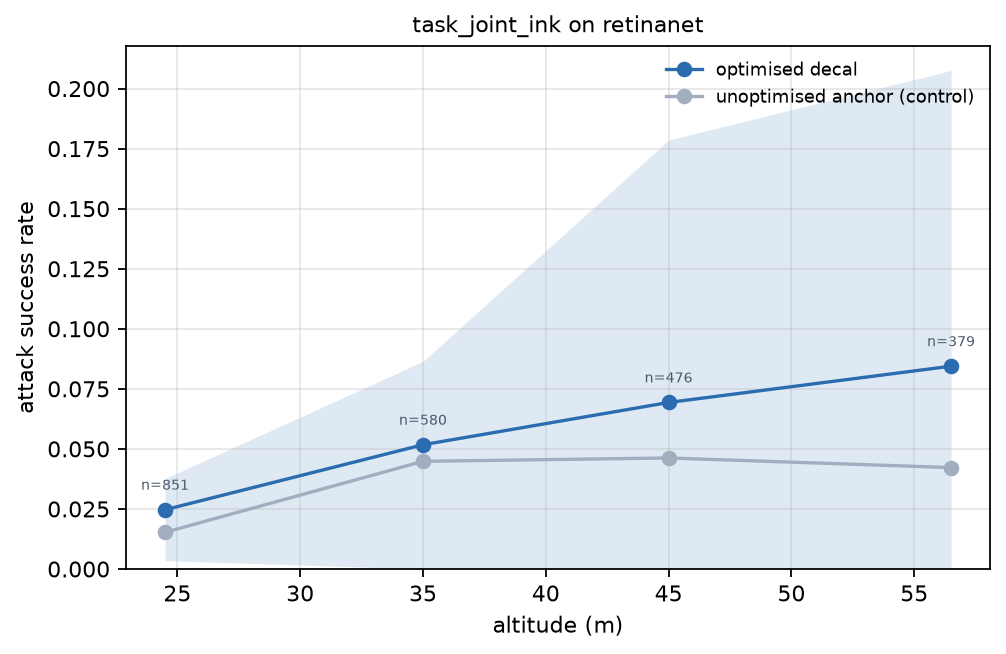}
    \caption{Altitude}
    \label{fig:margdet-alt}
  \end{subfigure}
  \caption{Attack success against each geometric axis for the measured-gamut
  patch on RetinaNet, with the size-matched control and vehicle-level confidence
  bands. The two curves remain close at every bin.}
  \label{fig:marg-det}
\end{figure}

\Cref{fig:marg-det} repeats the two geometric axes for the white-box detector.
There the attack and control curves lie close together at every bin---for
example $0.032$ against $0.026$ at \SI{15}{\degree} and $0.055$ against $0.054$
at \SI{65}{\degree}---and the confidence band on the attack curve contains the
control curve at every bin. This is the graphical form of the control
magnitudes reported in \Cref{sec:res-det}.

\subsection{Physical-world evaluation}\label{sec:res-phys}
The measured-gamut decal was printed, applied to two die-cast vehicles, and
photographed as 24 clean/patched pairs under three illumination conditions on a
matte, non-repeating substrate. Effective altitude was recovered per frame by
\Cref{eq:altrecover}; recovered effective altitudes span $11.6$ to
\SI{49.0}{\metre}, and 21 of 24 frames fall inside the simulated
$19$--$63$ metre range.

Of those 21 pairs, all 21 were attackable and none succeeded. Applying
\Cref{eq:cp} with $n = 21$ and $\alpha = 0.05$ gives an upper bound of
$p_{\max} = 0.133$.

The three excluded frames are reported rather than passed over. All three are
the closest standoff on vehicle~A, whose recovered altitude falls below the
simulated range. Two of the three were attackable and one of those two
succeeded, the only physical success observed in the session, with the vehicle
score falling from $0.570$ to $0.356$. It is excluded from the bound because
its altitude lies outside the range the bound is stated over, not because of
its outcome. Its direction agrees with the altitude dependence in
\Cref{tab:envelope}, where simulated \asr{} rises to $0.691$ in the lowest
altitude bin, but one success in two attackable frames supports no rate
estimate and none is computed. This does not exclude the simulated near-nadir rate of
$0.121$. The appropriate statement is therefore that the physical evaluation
bounds the transferred success rate at $0.133$ and did not detect transfer; it is
not powered to distinguish the absence of transfer from the simulated rate, and
the stronger claim is not asserted. The direction agrees with Hartnett et al.\
\cite{hartnett2022empirical} and Woo and Lee \cite{woo2026digital}.

The graded response does not alter this conclusion. Mean detection confidence
reduction over the in-range pairs is $+0.024$, with a cluster-bootstrap 95\%
interval of $[+0.001,+0.067]$ over the seven independent poses, which excludes
zero. That interval is not robust to removing a single pose: six of seven poses
have mean reductions between $-0.0002$ and $+0.0113$, while the seventh has
$+0.150$. Excluding it moves the mean to $+0.003$, and the median across poses is
$+0.0007$. The effect is therefore attributed to one pose and reported as an
anomaly rather than as a detected effect. Effective altitude and confidence
reduction are negatively related over the in-band range, but what is being
ordered is negligible. Pooled over the 21 pairs the correlation is
$r = -0.42$ across a $20.7$--$49.0$ metre span, and at the pose level, which
respects the clustering, $r = -0.43$ over the seven poses. The sign is stable
under leave-one-pose-out, where $r$ ranges from $-0.39$ to $-0.74$, and becomes
more negative rather than less when the anomalous pose above is removed. Its
direction agrees with the simulated altitude dependence in
\Cref{tab:envelope}. Two considerations bound what may be read into it. Among
the six poses other than the anomalous one, mean confidence reduction spans
only $-0.0002$ to $+0.0113$, so the correlation orders quantities that are
themselves close to zero. And altitude is partly confounded with vehicle
identity, the two models occupying different parts of the range at
$22.4$--$27.2$ and $20.7$--$48.1$ metres. No altitude threshold is claimed and
no rate is extrapolated from this trend.
%

\begin{figure}[htbp]
  \centering
  \begin{subfigure}[b]{0.48\linewidth}
    \includegraphics[width=\linewidth]{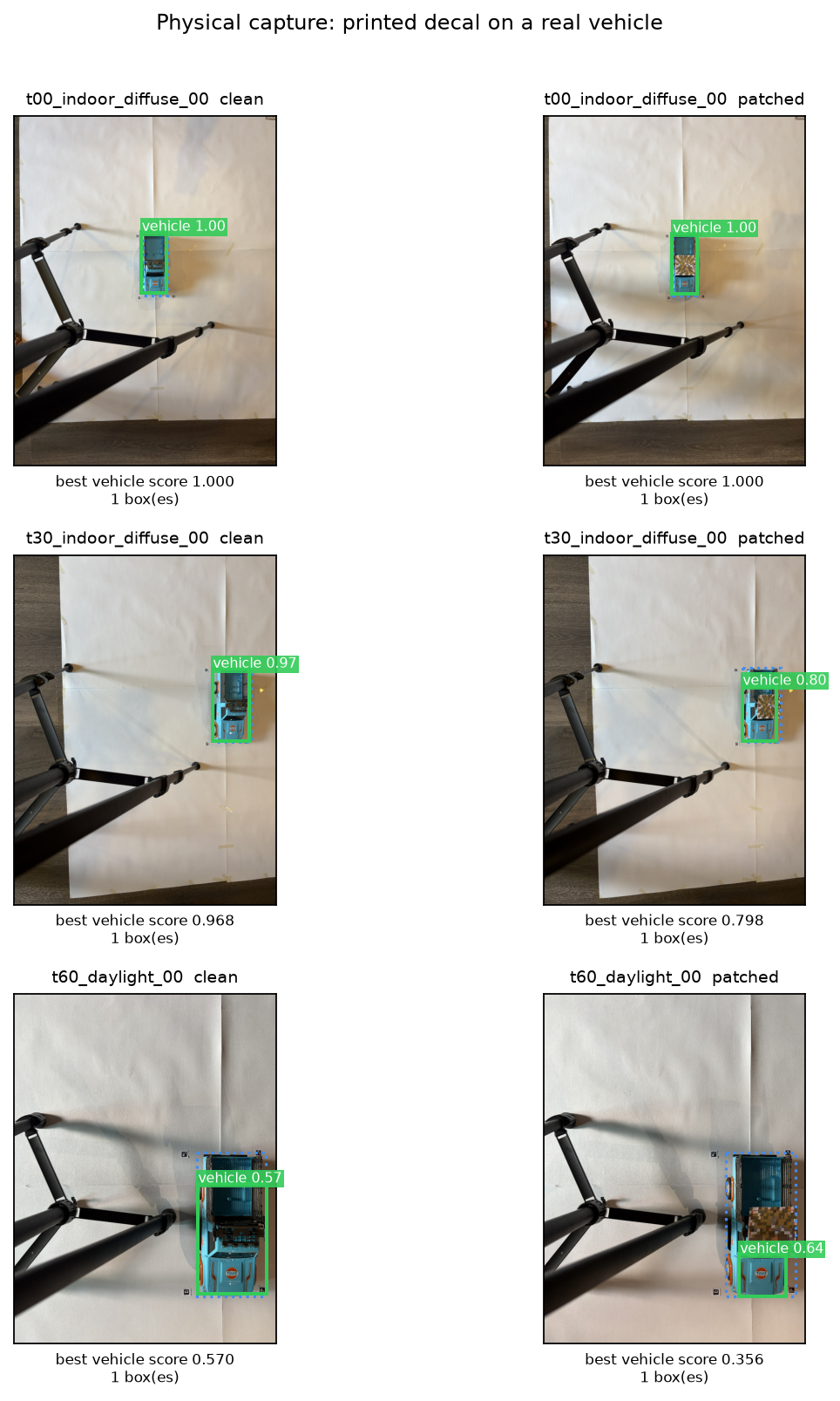}
    \caption{Vehicle A}
    \label{fig:phys-a}
  \end{subfigure}\hfill
  \begin{subfigure}[b]{0.48\linewidth}
    \includegraphics[width=\linewidth]{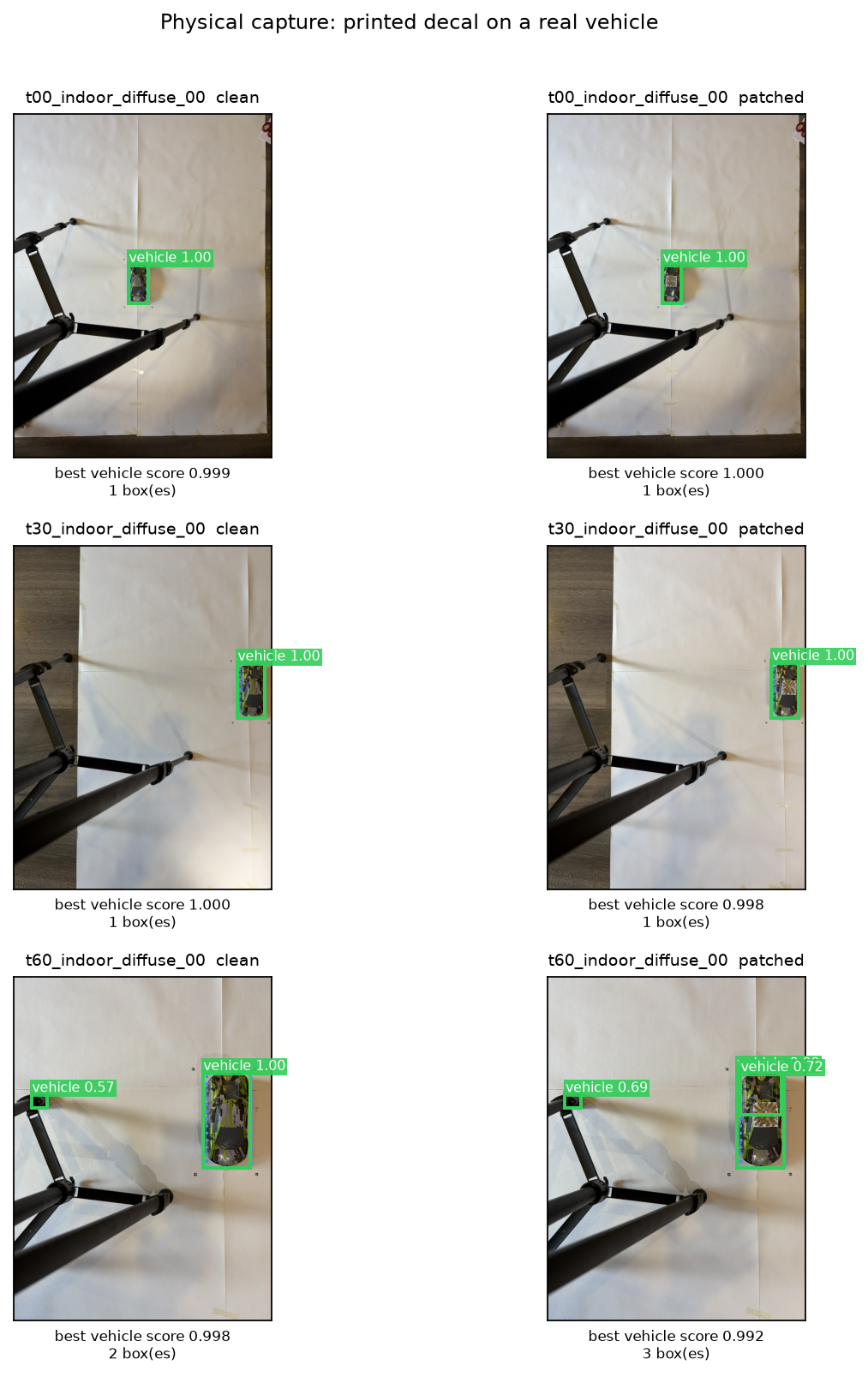}
    \caption{Vehicle B}
    \label{fig:phys-b}
  \end{subfigure}
  \caption{Physical captures with RetinaNet detections overlaid. Clean and
  patched frames are taken from the same pose seconds apart. Best vehicle
  confidence and the number of retained boxes are given beneath each panel.}
  \label{fig:phys}
\end{figure}

Clean detection succeeded in 23 of 24 frames, indicating that the capture
procedure was adequate and the victim located the vehicles reliably.
\Cref{fig:phys} shows representative clean and patched captures for both
vehicles with the victim's detections overlaid.

\FloatBarrier
\section{Discussion}\label{sec:discussion}
Three observations bear on how printed patch attacks are evaluated.

First, the printability constraint behaved as a regulariser rather than a
penalty on part of the segmentation arm, and \Cref{sec:res-seeds} shows the
effect exceeds run-to-run variation on three of six closed-set victims. It is
not a uniform property of those victims: two more are unresolved at three seeds
and one reverses sign. Whatever mechanism is at work is therefore
architecture-dependent, which any account of it has to accommodate.

\paragraph{Hypothesis H1 was tested and is not supported.} The account offered
in \Cref{sec:res-gamut} was that a palette a printer can reproduce is also a
less chromatic and lower-frequency one, and that the constrained optimiser
therefore spends its capacity on structure that survives imaging.
\Cref{sec:res-h1} tests it directly with eleven further patches and finds
against it. At matched cardinality a palette as chromatic as the cube attacks
as well as the measured one, so chromatic extent does not account for the
ordering. Varying cardinality from three inks to thirty-two produces no trend,
so the smaller search space does not account for it either. Both alternatives
named when the hypothesis was posed are therefore excluded.

What the experiment does show is that the arrangement of a palette matters more
than its summary statistics. Three random palettes matched on cardinality,
lightness and chroma span $0.035$ to $0.136$, a spread three times the
optimiser's own between-seed variation and larger than anything produced by
scaling chroma or changing cardinality. The measured gamut is evidently a
favourable set of colours, and not because it is small or muted.

We do not replace H1 with a second untested account. The result is reported as
a negative one: the mechanism behind \Cref{tab:net} remains open, two natural
explanations for it have been eliminated, and the evidence now points at the
identity of the colours rather than at any scalar summary of them. A mechanism
that survives this evidence would have to explain why particular hue
arrangements suit particular architectures, which \Cref{tab:seeds} shows is
itself victim-dependent.

Second, the detection arm does not distinguish the two ink sets, and the
magnitude of its controls indicates that much of what is commonly reported as
detection attack success in this setting is the effect of applying any object of
comparable size. This is a measurement issue rather than a property of the
attack, and is addressable by reporting size-matched controls.

These two observations also bear on how such attacks would be countered. The
decal is a localised, bounded region, which is the regime most patch defences
are built for: input preprocessing such as local gradient smoothing
\cite{naseer2019lgs}, saliency-based detection of the patch region
\cite{chou2020sentinet}, segment-and-inpaint pipelines for detectors
\cite{liu2022sac,jing2024pad}, and the certified constructions of Chiang et al.\
\cite{chiang2020certified} and Xiang et al.\ \cite{xiang2021patchguard}. Pathak
et al.\ \cite{pathak2024uavdefense} report a model-agnostic defence for the UAV
setting specifically. No defence was evaluated here, so this study says nothing
about whether the measured-gamut patch survives one.

Two features of the present results are relevant to anyone who does run that
evaluation. A defence evaluation in this setting needs the same size-matched
control, or the defence will be credited with suppressing an effect that is
partly the presence of the object rather than its content. And the realism
constraints reduce the patch's high-frequency content by construction
(\Cref{eq:texels}), so a defence that localises patches by their gradient
statistics \cite{naseer2019lgs} may find a band-limited, palette-constrained
decal harder to isolate than the unconstrained patches such defences are usually
tested against. Whether that holds is untested here and is stated as a
hypothesis, not a result.

Third, the relationship between the simulated and physical arms remains open.
Before drawing on it, the scope of the physical arm should be stated plainly,
because it is narrower than the term ``physical-world evaluation'' usually
implies. What was measured is printed decals on two die-cast models,
photographed indoors and on a driveway, on a pickup cargo bed rather than a cab
roof, at near-nadir only, with effective altitude recovered by scaling rather
than flown. What was \emph{not} measured is a full-size vehicle outdoors under
uncontrolled illumination, viewed from an airborne platform at off-nadir angles.
The bound reported below is a bound on the former. It is evidence about
sim-to-real transfer only to the extent that scale equivalence holds, which
\Cref{sec:method-phys} argues for coverage fraction, angular subtense and
ground sample distance but which does not extend to material response, surface
finish at full size, atmospheric path, or platform motion. No claim about
outdoor full-scale transfer is made anywhere in this paper, and the bound
should not be cited as one.

Within that scope, the simulated near-nadir rate of $0.121$ and the physical
upper bound of $0.133$ are mutually consistent, so these data do not establish
a sim-to-real gap; they establish that the physical sample is too small to
resolve one. That two
independent studies \cite{hartnett2022empirical,woo2026digital} report the same
direction makes a genuine gap more plausible, but none of the three has the
sample size to measure it.

A gap, if one exists, has candidate sources that these data cannot separate:
the renderer's material and illumination model, the printer and substrate, and
the difference between a die-cast model under indoor lighting and a vehicle
outdoors. Domain randomisation over rendering parameters is the standard
response where the reality gap has been characterised for other tasks
\cite{tobin2017domain,tremblay2018synthetic}, and the print-chain augmentation
of \Cref{sec:method-chain} is a narrow instance of the same idea applied to the
reproduction stage alone. Whether randomising the rendering parameters as well
would close the gap is untested here.

\FloatBarrier
\section{Limitations}\label{sec:limits}

\paragraph{Statistical power of the vehicle-level bootstrap.} All intervals are
bootstrapped over vehicles rather than instances. With three held-out vehicles, a
resample draws the same vehicle three times in one of every nine draws, and these
degenerate draws determine the lower tail. A 95\% vehicle-level interval can
therefore exclude zero only when an effect holds on every vehicle independently.
The phrase ``not distinguishable on three vehicles'' should be read as a
statement about the power of the test on this fleet, not as evidence of absence.

\paragraph{Three seeds resolve only large orderings.} \Cref{tab:seeds}
replicates every victim, so no ordering in this paper now rests on a single
run. Three seeds per condition remain few. An exact permutation test on three
against three admits ten splits and so cannot return a two-sided $p$ below
$0.10$, and the pooled standard deviation is itself estimated from two degrees
of freedom per condition. Four of the twelve victims have overlapping ranges
and are reported as unresolved rather than as null results: FCN-R50 and
UPerNet-ConvNeXt-T favour the measured gamut in the mean without separating,
RetinaNet does not separate, and Faster R-CNN separates by $0.0004$, which is
not a margin to rely on. A fifth, UPerNet-Swin-T, reverses sign relative to the
single run. More seeds would resolve these; they would not change the three
segmentation victims or the two open-vocabulary prompts that already
separate.

\paragraph{Universal patches act mostly through their footprint.} The
segmentation rates in this paper are those of universal decals, and
\Cref{sec:res-objcheck} shows that such decals remove only a few per cent of
vehicle pixels outside their own footprint, while the same objective fitted to
a single frame removes nearly half. A reader should therefore take the
segmentation rates as measuring a decal that suppresses what it covers and a
little beyond, not one that erases the vehicle. The per-frame comparison rests
on 24 instances of the two held-out pickups at nadir, on one victim, and does
not cover the box truck.

\paragraph{Single simulator and three held-out vehicles.} The corpus is synthetic
and rendered in one simulator. The physical evaluation is the intended
corrective, and as stated in \Cref{sec:res-phys} it bounds rather than
demonstrates transfer.

\paragraph{Scope of the physical evaluation.} All physical frames are near-nadir;
the capture labels index distance rather than viewing angle, so no conclusion is
drawn about off-nadir physical viewpoints. The decal was applied to a pickup
cargo bed rather than a cab roof. The bed is a flat panel an operator could use,
and its coverage fraction of $0.722$ is comparable to the simulated box-truck
roof's $0.776$, but it is not the panel the optimiser targeted. A roof-mounted
decal on a box body was not tested.

\paragraph{Open-vocabulary evaluation is control-dominated.} The CLIPSeg control
alone reaches $0.326$--$0.370$, indicating that an unoptimised cargo marking
defeats this victim on approximately one third of instances. Reporting the raw
rates of $0.42$--$0.56$ without the control would overstate the contribution of
the optimisation by roughly a factor of three. One prompt, ``a vehicle seen from
above'', has an empty attackable subset under both patches and is undefined.

\paragraph{One victim is not comparable to the others.} DeepLabv3-R101 shows the
largest nets in \Cref{tab:net} but locates only 678 of 2534 instances cleanly
(27 per cent), against 85--88 per cent for FCN-R50 and SegFormer-B2. A net of
$+0.596$ over a quarter of the fleet is not equivalent to the same value over all
of it.

\paragraph{Segmentation could not be evaluated physically, and the reason is
informative.} Only 5 of 24 clean physical frames cleared the $0.50$
attackability threshold, against $1800$ of $2534$ in simulation for the same
victim, so the physical segmentation rate is undefined rather than zero and no
value is reported.

The shortfall is not uniform across the capture. Clean segmentation IoU falls
monotonically with the standoff label, from $0.49$--$0.81$ at the furthest
standoff, where 5 of 6 frames clear the gate, through $0.23$--$0.50$ and
$0.01$--$0.18$, to $0.000$--$0.020$ at the closest, where none do. The detector
is largely unaffected over the same frames: clean vehicle score is at least
$0.95$ in 20 of 24. The failure is therefore specific to dense mask prediction
rather than to the capture or the subject. The victim still locates the
vehicle; the mask it returns overlaps the annotation too little for an attack
test to be defined.

Two explanations are consistent with this and the data do not separate them. At
close standoff the die-cast model subtends far more of the frame than any
vehicle in the training corpus, so the failure may be one of scale. It may
instead be appearance, the proportions, materials and specular response of a
die-cast model differing from a rendered vehicle in ways a dense decoder is
more sensitive to than a box regressor. Effective altitude alone does not
account for it: vehicle~B at $44.7$ and \SI{36.0}{\metre}, both inside the
simulated range, returns IoU of $0.23$ and $0.01$, while vehicle~A at
\SI{24.7}{\metre} returns $0.38$--$0.50$. Any future scaled capture intended to
test segmentation should verify the clean-frame gate before the session rather
than after it, which this session did not.

\FloatBarrier
\section{Future Directions}\label{sec:future}
Four extensions follow from the limitations, in decreasing order of how much
they would change what can be claimed.

Explaining the palette effect is the most consequential, and
\Cref{sec:res-h1} has narrowed rather than closed it. Chromatic extent and
cardinality are excluded; what remains is that particular colours suit
particular victims. The natural next experiment follows the variance: since
between-palette variation exceeds between-seed variation threefold, a larger
sample of random palettes at matched statistics, each replicated over seeds,
would establish the distribution of attack success over palettes and show
whether the measured gamut is an outlier in it or merely a good draw. Relating
the outcome to where a palette's colours fall relative to the victim's own
colour sensitivity would then be the first candidate mechanism worth testing.

A physical sample sufficient to resolve a rate of $0.121$ against a null of
zero would require on the order of $10^{2}$ attackable pairs at fixed geometry,
which is achievable with a fixed rig; extending that rig to a full-size vehicle
outdoors would address the scope limit set out in \Cref{sec:discussion}, which
the present scaled capture does not reach.

A held-out fleet larger than three vehicles would restore the power of the
vehicle-level bootstrap. As \Cref{sec:method-metrics} notes, this is a
rendering cost rather than a change of method: it requires additional vehicle
models and a re-render, after which intervals would stop being dominated by
degenerate resamples.

Finally, seed replication could be extended beyond the roster reported in
\Cref{tab:seeds} to the optimisation levers of \Cref{sec:res-realism}, whose
two-point comparisons rest on single runs.

\FloatBarrier
\section{Conclusions}\label{sec:conclusion}
Printability is generally treated as a constraint that reduces adversarial patch
effectiveness. When the constraint was measured rather than assumed and imposed
on an otherwise identical optimisation, that expectation failed on part of the
segmentation arm and held elsewhere. Across three seeds per condition and the
full victim roster, the measured-gamut patch achieved the higher net attack
success on three of six closed-set segmentation victims with disjoint seed
ranges, was consistent with doing so on two more, and reversed sign within
noise on the sixth. On the open-vocabulary segmenter and on two of four
detectors the colour-cube patch was reliably the stronger, and no detector
reached a net above $+0.026$ under either ink set. The useful statement is
therefore not that printability helps or hurts, but that its sign depends on the
victim and must be measured per architecture rather than assumed.

The obvious interpretation of that result---that a reproducible palette is also
a less chromatic, lower-frequency one---was tested with eleven further patches
and does not survive. Neither chromatic extent nor palette cardinality accounts
for the ordering, while palettes matched on both but differing in hue placement
span a range three times the optimiser's own seed variation. What a printer's
gamut contributes is a particular set of colours, not a smaller or duller one,
and the mechanism behind that is left open rather than replaced with a second
untested account. Grad-CAM attributions indicate the attack operates by
suppressing vehicle evidence beneath the decal.

The physical evaluation did not detect transfer and bounds the transferred
success rate at $0.133$ with 95\% confidence, an interval that still contains the
simulated near-nadir rate. A bound is therefore reported rather than a
demonstration, in a direction consistent with two independent prior studies.
Separately, the magnitude of the size-matched controls indicates that most raw
detection attack success in this setting is attributable to the presence of an
applied object rather than to its optimised content. These results narrow what
can presently be claimed for printed adversarial decals in aerial perception.

\section*{Reproducibility Statement}
All results derive from a single 2534-instance holdout. Each optimisation run
archives its configuration, seed, loss history and realism metrics. The
alternative attack terms and the level-4 reparametrisation of
\Cref{sec:res-objcheck} are configuration options; the defaults reproduce the
reported patches. Four defects
identified while constructing the physical pipeline are documented in the
accompanying repository; each returned a plausible incorrect value rather than
raising an error, namely an incorrect perspective-$n$-point solver selection for
the three-marker case, an EXIF sub-IFD read that silently substituted an assumed
field of view, a capture plan reporting a decal size taken from configuration
rather than from the patch, and default fiducial detector parameters that fail at
the marker size the procedure computes.


\section*{Author Contributions}

\section*{Funding}

\section*{Ethical Statement}
This study involves no human participants, no personally identifying data
and no animals. The physical evaluation was carried out on die-cast model
vehicles on private property.

\section*{Data Availability Statement}
The synthetic corpus, the measured printer gamut, every optimised patch with
its configuration and loss history, the per-instance evaluation records
underlying all reported rates, and the physical capture set are archived in the
accompanying repository. The palette families of
\Cref{sec:method-palettes} are regenerated by \texttt{rp\_palettes.py} and the
altitude recovery of \Cref{sec:method-phys} by \texttt{rp\_altitude.py}, so
both are reproducible from the released code rather than only reported.

\section*{Acknowledgements}

\section*{Conflicts of Interest}
The authors declare no conflict of interest.

\bibliographystyle{unsrt}
\bibliography{references}

\end{document}